\documentclass[aps,preprintnumbers,showpacs,amsmath,amssymb,floatfix,10pt,prd,superscriptaddress,nofootinbib,twocolumn]{revtex4-2}
\usepackage[utf8]{inputenc}
\usepackage{float}
\usepackage{subcaption}
\usepackage{hyperref}
\hypersetup{
	colorlinks=true,
	linkcolor=blue,
	filecolor=magenta,
	urlcolor=magenta,
}
\usepackage{mathtools}
\usepackage{amssymb}
\usepackage{amsmath}
\usepackage{graphicx}
\usepackage{latexsym}
\usepackage{array}
\usepackage{orcidlink}

\begin{document}
\title{Physics Informed Kolmogorov-Arnold Neural Networks for Dynamical Analysis via Efficent-KAN and WAV-KAN}

\author{Subhajit Patra\orcidlink{0009-0001-6043-8747}}
\email{eey247540@iitd.ac.in}
\affiliation{Department of Electrical Engineering, Indian Institute of Technology, Delhi, India}

\author{Sonali Panda \orcidlink{0009-0008-6027-8342}}
\email{sp.sonalpanda@gmail.com}
\affiliation{Department of Physics, Indian Institute of Technology,  Dhanbad, India}

\author{Bikram Keshari Parida \orcidlink{0000-0003-1204-357X}}
\email{parida.bikram90.bkp@gmail.com}
\affiliation{Artificial Intelligence and Image Processing Lab., Sun Moon University, South Korea}

\author{Mahima Arya \orcidlink{0000-0002-1847-9705}}
\email{aryamahima@gmail.com}
\affiliation{School of Artificial Intelligence, Amrita Vishwa Vidyapeetham, Coimbatore 641112, India}

\author{Kurt Jacobs\orcidlink{0000-0003-0828-6421}}
\email{dr.kurt.jacobs@gmail.com}
\affiliation{United States DEVCOM Army Research Laboratory, Adelphi, Maryland 20783, USA}
\affiliation{Department of Physics, University of Massachusetts at Boston, Boston, Massachusetts 02125, USA}

\author{Denys I. Bondar \orcidlink{0000-0002-3626-4804}}
\email{dbondar@tulane.edu}
\affiliation{Department of Physics and Engineering Physics, Tulane University, New Orleans, LA 70118, USA}

\author{Abhijit Sen \orcidlink{0000-0003-2783-1763}}
\email{asen1@tulane.edu}
\affiliation{Department of Physics and Engineering Physics, Tulane University, New Orleans, LA 70118, USA}

\begin{abstract}
Physics-informed neural networks have proven to be a powerful tool for solving differential equations, leveraging the principles of physics to inform the learning process. However, traditional deep neural networks often face challenges in achieving high accuracy without incurring significant computational costs. In this work, we implement the Physics-Informed Kolmogorov-Arnold Neural Networks (PIKAN) through efficient-KAN and WAV-KAN, which utilize the Kolmogorov-Arnold representation theorem. PIKAN demonstrates superior performance compared to conventional deep neural networks, achieving the same level of accuracy with fewer layers and reduced computational overhead. We explore both B-spline and wavelet-based implementations of PIKAN and benchmark their performance across various ordinary and partial differential equations using unsupervised (data-free) and supervised (data-driven) techniques. For certain differential equations, the data-free approach suffices to find accurate solutions, while in more complex scenarios, the data-driven method enhances the PIKAN's ability to converge to the correct solution. We validate our results against numerical solutions and achieve $99 \%$ accuracy in most scenarios.
\end{abstract}

\maketitle

\section{Introduction}

The advent of deep learning and its use cases in solving complicated tasks related to computer vision, natural language processing, speech, \emph{etc.}, has led to state-of-the-art applications in industries like healthcare, finance, robotics, to name a few. Further, using deep neural networks (DNNs) in solving differential equations through Physics Informed Neural Networks (PINNs) is another breakthrough that offered a new framework for solving partial differential equations \cite{PINN1}. Since then the field of PINN has received a lot of attention (e.g., see review \cite{PINNReview}) and is extended to solve fractional equations, integral-differential equations, and stochastic partial differential equations \cite{PINNother1, PINNother2, PINNother3}. PINN has been developed to be more robust and accurate \cite{PINNremedy1} because the original form of PINN has drawbacks \cite{PINNdrawback, PINNdrawback1, PINNdrawback2, PINNdrawback3, PINNdrawback4, PINNdrawback5}, which are emanate from deep networks.

Recently, a promising alternative to the traditional multilayer perceptron has been proposed: the Kolmogorov-Arnold Neural Network (KAN) \cite{KAN}. While the universal approximation theorem \cite{UAT} is foundational to deep learning, KAN is based on the Kolmogorov-Arnold representation theorem. This theorem establishes that any multivariate continuous function can be decomposed into sums of univariate continuous functions. KANs leverage the power of splines and multi-layer perceptrons, as KAN is essentially a combination of the two. KAN offers a distinct advantage over traditional Multi-layer Perceptrons (MLPs) by incorporating learnable activation functions in addition to learnable weights \cite{KAN}. This characteristic enhances the ability of KANs to approximate solutions to differential equations with greater accuracy and fidelity compared to conventional neural networks (e.g., PINNs). By allowing the activation functions themselves to adapt during training, KANs can better capture the intricate dynamics and behaviors inherent in differential equations, thereby producing solutions that closely align with actual solutions.

A KAN-based physics-informed neural network (PIKAN) has been shown to solve the 2D Poisson equation~\cite{KAN} for a variety of geometries as well as other partial differential equations~\cite{wang_kolmogorov_2024}. KAN was also successfully applied to system identification~\cite{koenig_kan-odes_2024}. A thorough comparison of PIKAN based on different variations of KAN have been made in~\cite{PIKAN1}. However, more recent implementations of KAN, such as WAV-KAN~\cite{WAV-KAN} or efficient-KAN~\cite{Blealtan2024}, have not been utilized for solving differential equations.

In this article, we present a PIKAN analysis of various differential equations via efficient-KAN~\cite{Blealtan2024} and WAV-KAN~\cite{WAV-KAN}. We explore the data-free approach -- Data-Free PIKANs (DF-PIKANs), which solve equations without relying on external data, making them ideal for problems where data is scarce or unavailable. Also, the data-driven approach is utilized -- Data-Driven PIKANs (DD-PIKANs), which uses existing datasets to enhance solution accuracy. The use of data provides guidance during training, helping the model to learn more accurately. We will demonstrate the performance of both approaches, drawing parallels to similar concepts  PINNs. Note that the key advantage of using PIKANs is that they require less experimentation with different architectures. This is because PIKANs (via efficient-KAN and WAV-KAN) can effectively capture the solutions of differential equations with simpler, lower-complexity architectures.

The rest is organized as follows: In Section \ref{KAN_int}, we provide an introduction to KAN, discussing its theoretical foundation and advantages over traditional neural networks. Section \ref{pikan_imp} focuses on the implementation details of the PIKAN, explaining the construction and training process of PIKANs, including both B-spline and wavelet-based approaches.  Sections~\ref{SecODESingle}, \ref{COUPLED_DEQN}, \ref{osc_dyn} and \ref{npde} present a series of case studies that showcase the effectiveness of PIKANs in achieving high accuracy across different types of differential equations, validated against numerical solutions. Finally, in Section \ref{concl}, we summarize our findings, highlighting the benefits and potential future directions for PIKANs in the context of differential equations analysis.

\section{Introduction to Kolmogorov Arnold Networks (KAN)} \label{KAN_int}

Recall that DNNs is based on the Universal Approximation Theorem, which reads as follows: 
Let $\sigma$ be any continuous sigmoidal function. Then finite sums of the form
$$
g(\textbf{x})=\sum_{j=1}^N \alpha_j \sigma\left(\textbf{w}_j^T \textbf{x}+b_j\right), \text { where } \textbf{x}, \textbf{w}_j \in \mathbb{R}^n, \alpha_j, b_j \in \mathbb{R}
$$
are dense in the set of continius functions on $[0, 1]^n$ \cite{UAT}.
Thus, for any continuous function $f(\textbf{x})$, there is a sum $g(\textbf{x})$ of the above form such that
$$
|g(\textbf{x})-f(\textbf{x})|<\varepsilon.
$$
In neural network context, this theorem shows that a sufficiently smooth function $f(\textbf{x})$ can be approximated arbitrarily closely on a compact set using a two-layer neural network (NN) with appropriate weights and activation functions \cite{UAT1,UAT2}. A two layer NN is a network that consists of an input layer, one hidden layer, and an output layer. The hidden layer has neurons with activation functions, and the output layer typically performs a weighted sum of the hidden layer outputs.\\

KAN is based on the Kolmogorov-Arnold representation theorem  stating that any continuous function of multiple variables can be represented as a superposition of continuous functions of a single variable and addition \cite{KAT1,KAT2, KAT3}. Formally, it states that,
given a continuous function $f:[0,1]^n \rightarrow \mathbb{R}$, there exist uni-variate continuous functions $\phi_{i, j}$ and $\psi_i$ such that
$$
f(\textbf{x}) = f\left(x_1, x_2, \ldots, x_n\right)=\sum_{i=0}^{2 n + 1 } \psi_i\left(\sum_{j=1}^n \phi_{i, j}\left(x_j\right)\right).
$$

Note that the potential of utilizing KAN for constructing neural networks has been attempted  previously ~\cite{FKAT1,FKAT2,FKAT3,FKAT4,FKAT5,FKAT6}. However, in \cite{KAN}, the shortcomings of the previous attempts were finally resolved, resulting in the comprehensive construction of neural networks (KANs) utilizing the Kolmogorov-Arnold theorem. In KANs, nodes perform the function of summing incoming signals without introducing any non-linear transformations. Conversely, edges within the network incorporate learnable activation functions, which are weighted combination of basis function and a B-splines. Throughout the training process, KANs optimize these spline activation functions to match the target function. Further, the learning parameters are fundamentally different from those in traditional neural networks, which use weights and biases. In KANs, the primary parameters are the coefficients of the learnable uni-variate activation functions. 

In this paper, we will use efficient-KAN (for code implementation refer to \cite{Blealtan2024}), a reformulation of originally proposed KAN \cite{KAN} which significantly reduces the memory cost and make computation faster. Further, we also utilize an alternative to KAN (or efficiemt-KAN) called as Wavelet Kolmogorov-Arnold Networks (WAV-KAN) ~\cite{WAV-KAN} which uses wavelets instead of B-splines. 

In this paper, we will leverage both the use of efficient-KAN and WAV-KAN in solving various differential equations. Note that we will use the name PIKAN consistently, regardless of the type of KAN being used. However, we will make it clear if we use the efficient-KAN or WAV-KAN for solving the differential equation.

\section{Implementation of PIKAN} \label{pikan_imp}

The implementation of PIKAN (either with efficient KAN or WAV-KAN) is similar to that of PINN except for the fact that instead of using DNNs, one uses KANs. For simplicity, we will define it for an ordinary differential equation and introduce DF-PIKAN. For the differential equation subject to the initial condition as follows:
\begin{equation}
    \frac{dy(x)}{dx} - f(y(x),x) = 0 , \qquad y(a) = y_{a}
\end{equation}
where $x\in[0,1]$, the aim is to train a neural network which finds a function $\hat{y}_{\theta}(x)$ such that satisfies the equation in an approximate manner, i.e.,
\begin{equation}
    \frac{d\hat{y}_{\theta}(x)}{dx} - f(\hat{y}_{\theta}(x),x) \approx 0, \qquad \hat{y}_{\theta}(a) = y_{a}
    \label{eqy}
\end{equation}
within the specified domain of the original differential equation. Note that $\theta$ denotes the tunable parameters (e.g, weights, biases) of neural networks. In KANs, these parameters are the coefficients of the learnable uni-variate activation functions applied to each input. In order to achieve the approximate accurate form $\hat{y}_{\theta}(x)$, the main trick is to introduce the correct loss function for the neural network. Let us define the residual of the differential equation as
\begin{equation}
    \mathcal{R}_\theta(x_{\mathrm{r}}^i)=\frac{d\hat{y}_{\theta}(x_{\mathrm{r}}^{i})}{dx} - f(\hat{y}_{\theta}( x_{\mathrm{r}}^{i}), x_{\mathrm{r}}^{i})
\end{equation}
where $\{ x_{r}^{i}\}_{i=1}^{N_{r} }$ represents specific points in the domain of the variable 
$x$. These points are either vertices of a fixed mesh or points that are randomly sampled during each iteration of a gradient descent algorithm. Now using the residual loss defined above, physics loss is defined as:
\begin{equation}
    \mathcal{L}_{\mathrm{r}}=\frac{1}{N_{r}}\sum_{i=1}^{N_{r}}\left|\mathcal{R}_\theta(x_{\mathrm{r}}^{i})\right|^2.
\end{equation}
To ensure that the right-hand side (RHS) of Eq \ref{eqy} is approximately zero, we need to minimize the physics loss. This involves reducing the discrepancy between the derivative of the neural network output \(\frac{d\hat{y}_{\theta}(x)}{dx}\) and the function \(f(\hat{y}_{\theta}(x), x)\) at the mesh points \(x_r^i\). By making the physics loss as small as possible, we ensure that the neural network solution adheres closely to the underlying physical laws represented by the differential equation.

However, we have not yet considered how to incorporate the information about boundary condition/initial conditions. This can be done as either hard or soft constraints. In the hard constraint style, to incorporate the initial condition $y(a) = y_{0}$, the output of the neural network is modified such that $\hat{y}_{\theta}(x)$ $\mapsto$ $y_{a} +(x-a)  \hat{y}_{\theta}(x)$. In such a case, the physics loss is the only term in the loss function. In the soft constraint style, the initial condition information is included in the loss function. Thus, including boundary condition as a soft constraint, the final loss function for the neural network becomes
\begin{equation}
\begin{split}
    \mathcal{L}_{\mathrm{final}} & = \mathcal{L}_{\mathrm{r}} + \mathcal{L}_{\mathrm{bc}}  \\
    & =\frac{1}{N_{r}}\sum_{i=1}^{N_{r}}\left|\mathcal{R}_\theta(x_{\mathrm{r}}^{i})\right|^2 + \frac{1}{N_{bc}}\sum_{i=1}^{N_{bc}}\left|\hat{y}_{\theta}(x_{\mathrm{bc}}^{i})-y_{a}^i   \right|^{2},
\end{split}
\end{equation}

Where $ N_{r} $ is the maximum number of collocation points used to evaluate the  residuals and $ N_{bc} $ is the maximum number of boundary collocation points. In our example, $N_{bc} = 1$, since we have only one boundary condition.

However, as straightforward as the formulation may appear, it is not without subtleties. First, note that different terms in the loss function can have different magnitudes and, therefore, some terms might dominate the training process, while others are neglected. In order to ensure that all parts of the loss contribute appropriately to the training process, the weights are often multiplied in the loss function to balance different terms. In such a case, the loss function with multiplicative weight (also including initial condition if any) takes the following form
\begin{equation}
    \mathcal{L}_{\mathrm{final}}  = \lambda_{\mathrm{r}}\mathcal{L}_{\mathrm{r}} + \lambda_{\mathrm{bc}} \mathcal{L}_{\mathrm{bc}} +  \lambda_{\mathrm{ic}} \mathcal{L}_{\mathrm{ic}} 
    \label{weight-loss}
\end{equation}
The DF-PIKANs utilize the loss function as in Eq (\ref{weight-loss}). Note that
\begin{equation}
    \mathcal{L}_{\mathrm{final}}^{DF}  =  \mathcal{L}_{\mathrm{final}}
    \label{weight-loss-DF}
\end{equation}
where the superscript $DF$ in $\mathcal{L}_{\mathrm{final}}^{DF} $ denotes that it pertains to a Data-Free approach. 

In this section, we will now delve into DD-PIKANs (similar to DD-PINNs \cite{PINN1, dfdd1}) where the loss function Eq (\ref{weight-loss}) is modified to include observational or simulated data. In other words, DF-PIKNs rely on governing physical equations for learning, while DD-PIKANs use additional data in the loss function to improve accuracy and robustness during training. The loss function is given by
\begin{equation}
    \mathcal{L}_{\mathrm{final}}^{DD}  = \lambda_{\mathrm{r}}\mathcal{L}_{\mathrm{r}} + \lambda_{\mathrm{bc}} \mathcal{L}_{\mathrm{bc}} +  \lambda_{\mathrm{ic}} \mathcal{L}_{\mathrm{ic}} + \lambda_{\mathrm{data}} \mathcal{L}_{\mathrm{data}} 
    \label{weight-loss-DD}
\end{equation}
where 
\begin{equation}
    \mathcal{L}_{\mathrm{data}}=\frac{1}{N_{data}} \sum_{i=1}^{N_{data}}\left|\hat{u}_\theta \left(x_{\mathrm{data}}^{i}\right)-u\left(x_{\mathrm{data}}^{i}\right)\right|^2
\end{equation}
which ensures the solution fits the observed collocation data points $\left(x_{\mathrm{data}}^{i}, u\left(x_{\mathrm{data}}^{i}\right)\right)$. \\

For completeness, we write the DD-PIKAN loss function for differential equation with two variables, for example, Burger equation.
Burgers' equation in one spatial dimension $x$ and time $t$ is given by:
$$
\frac{\partial u}{\partial t}+u \frac{\partial u}{\partial x}=\nu \frac{\partial^2 u}{\partial x^2}
$$
where $u=u(x, t)$ is the velocity field, and $\nu$ is the kinematic viscosity -- a given parameter. The initial condition specifying the value of $u$ at the initial time $t=0$ is given by
$$
u(x, 0)=u_0(x)
$$
and the boundary condition that specifies the behavior of $u$ at the boundaries of the spatial domain $x$, often $x=0$ and $x=L$ :
$$
u(0, t)=u_L(t), \quad u(L, t)=u_R(t)
$$
The residual term,  physics loss, initial condition loss, boundary loss and the total loss are the following
\begin{align}
     R_{\theta}(x,t) & = \frac{\partial u}{\partial t}+u \frac{\partial u}{\partial x} - \nu \frac{\partial^2 u}{\partial x^2},  
    \nonumber
    \\
    \mathcal{L}_{\mathrm{r}} &=  \frac{1}{N_{r}} \sum_{i=1}^{N_{r}} |\mathcal{R}_\theta(x_{\mathrm{r}}^{i},t_{\mathrm{r}}^{i})|^2, \nonumber\\ 
    \mathcal{L}_{\mathrm{ic}} &= \frac{1}{N_{ic}} \sum_{j=1}^{N_{ic}} \left( \hat{u}(x_{\mathrm{ic}}^{j}, 0) - u_{0}(x_{\mathrm{ic}}^{j}) \right)^2, \nonumber
    \\
    S_{1} & = \frac{1}{N_{bc}} \sum_{k=1}^{N_{bc}} \left( \hat{u}(0, t_{\mathrm{bc}}^{k}) -  u_{L}(t_{\mathrm{bc}}^{k})\right)^2, \nonumber
    \\
    S_{2} & = \frac{1}{N_{bc}} \sum_{k=1}^{N_{bc}} \left( \hat{u}(L, t_{\mathrm{bc}}^{k})  - u_{R}(t_{\mathrm{bc}}^{k})\right)^2, \nonumber
    \\
    \mathcal{L}_{bc} &= S_{1} + S_{2}, \nonumber
    \\
    \mathcal{L}_{\mathrm{final}}^{DF} &=  \mathcal{L}_{\mathrm{r}} +  \mathcal{L}_{\mathrm{ic}} + \mathcal{L}_{\mathrm{bc}}.
\end{align}

We note that it is well-known that the data-free approach of PINNs presents some limitations in finding a correct solution to many differential equations. For example, for one-dimensional convection problems, PINNs achieve a good solution only for small values of convection coefficients  \cite{NEURIPS1, dfdd1,dfdd2,dfd3}. In scenarios like this, adopting a data-driven approach yield a significant improvement. By incorporating a data loss term, the training process is guided more effectively toward achieving correct solutions . We will demonstrate that the same applies to the Burger equation below.

\section{Linear ordinary differential equation}\label{SecODESingle}

Consider the following initial value problem:
\begin{equation}\label{EqODE11}
    \frac{dy}{dx} = 3x^2, \qquad y(0) = 1 . 
\end{equation}
Let us define the residual loss function as
\begin{equation*}
    \mathcal{R}_\theta(x) = \frac{d\hat{y}_{\theta}(x)}{dx} - 3x^2, 
\end{equation*}
where $\hat{y}_{\theta}(x)$ is the PIKAN solution which we intend to learn. From Eq (\ref{weight-loss}) and Eq (\ref{weight-loss-DF}), the final loss function reads as
\begin{equation}\label{EqLDFODE}
    \mathcal{L}^{DF}_\mathrm{final} = \frac{1}{N_{r}} \sum_{i=1}^{N_{r}} \left| \mathcal{R}_\theta(x_{\mathrm{r}}^{i}) \right| ^2 +  \left| \hat{y}_{\theta}(0) - 1 \right|^2 
\end{equation}
where $x_{\mathrm{r}}^{i}$ are the collocation points with the weights $\lambda_{\mathrm{r}} = \lambda_{\mathrm{ic}} = 1$ and $\lambda_{\mathrm{bc}} = 0$. \\

\begin{figure}
    \centering
    \begin{subfigure}[t]{0.5\textwidth}
        \centering
        \includegraphics[width=\textwidth]{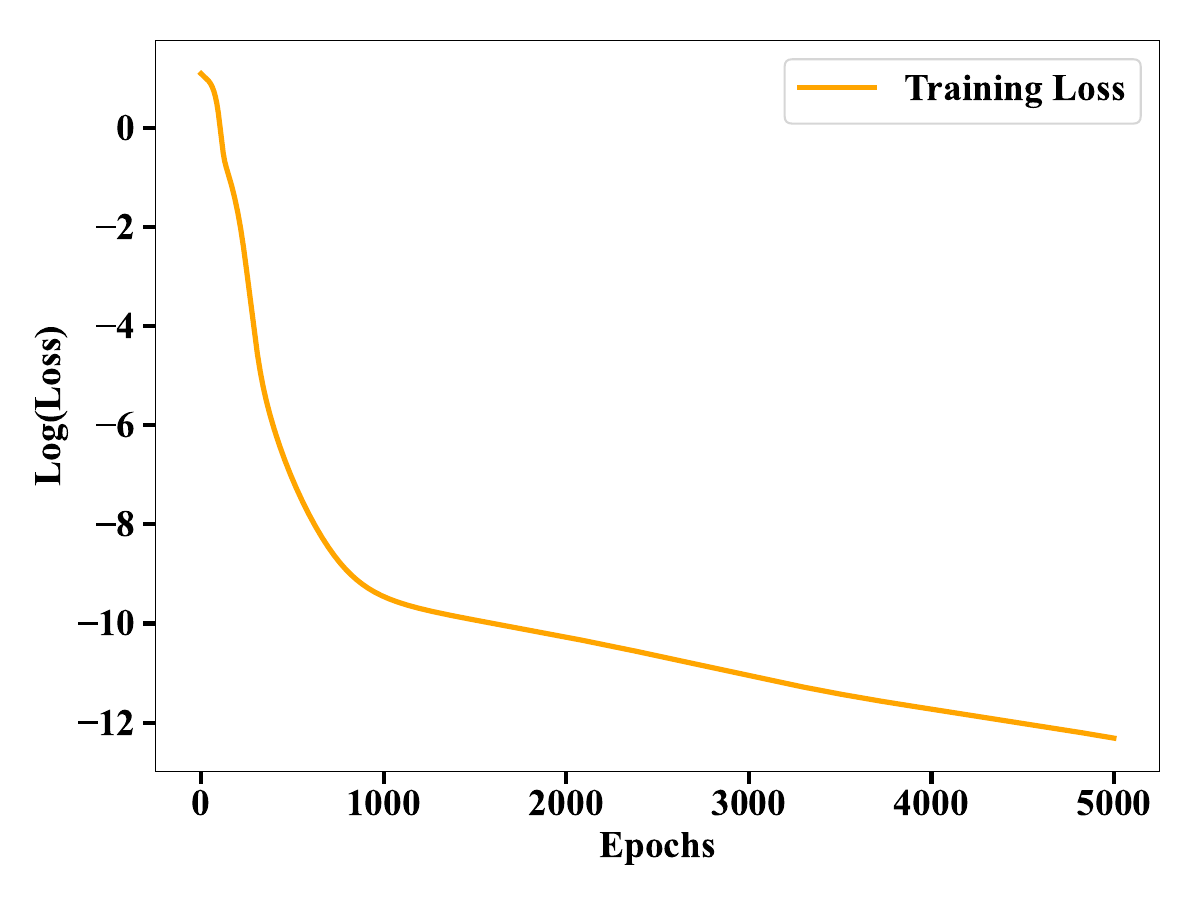}
        \caption{  }
        \label{fig-LDEa}
    \end{subfigure}%
    \hfill
    \begin{subfigure}[t]{0.5\textwidth}
        \centering
        \includegraphics[width=\textwidth]{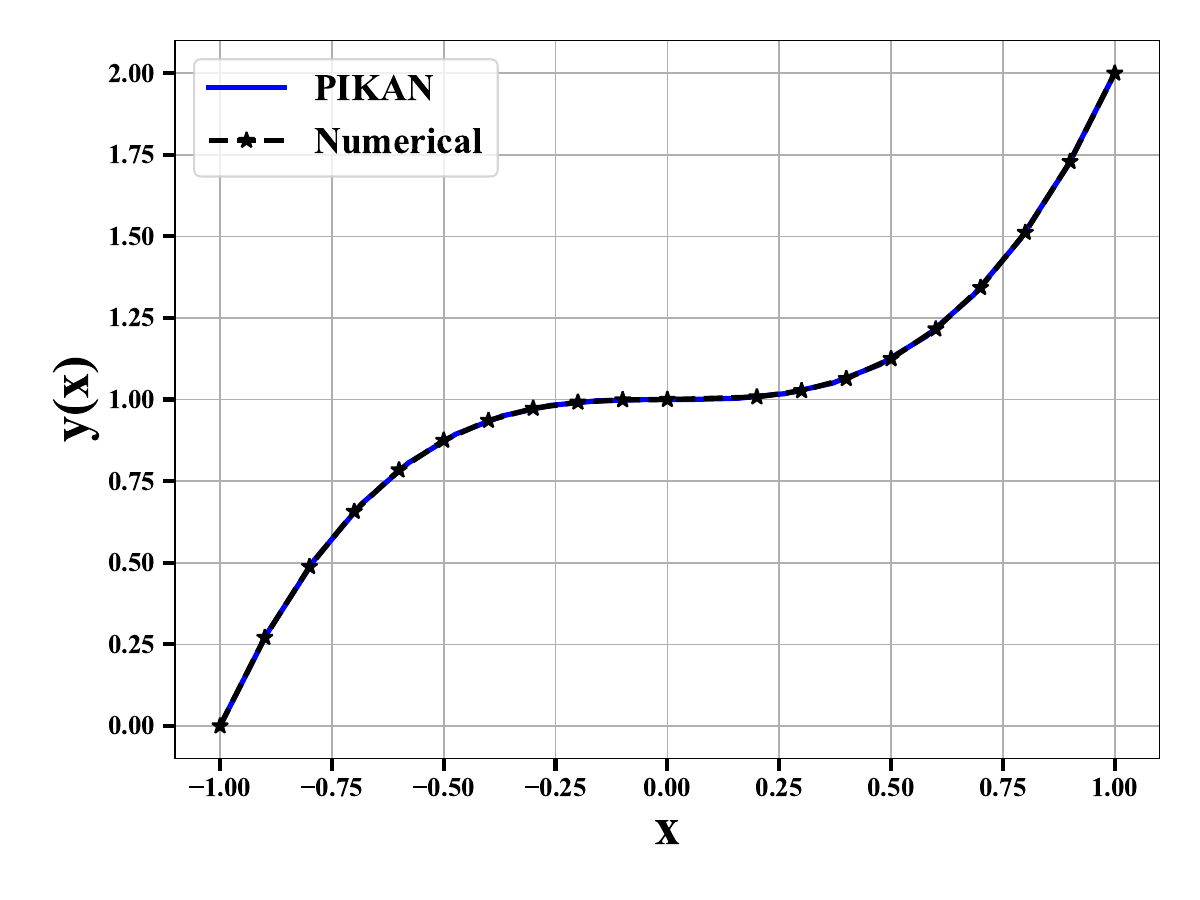}
        \caption{ }
        \label{fig-LDEb}
    \end{subfigure}
    \caption{(a) Illustrates the relationship between loss~\eqref{EqLDFODE} and epoch. (b) A comparison between the PIKAN predicted solution and numerically exact solution of Eq.~\eqref{EqODE11}.}\label{fig-LDE}
\end{figure}

We utilized the efficient-KAN method to solve the differential equation, employing  collocation points $N_{x} = 100$ equally distributed on the interval $-1 \leq x \leq  1$. The model was trained for $5000$ epochs with the learning rate $\eta = 0.001$. The base activation function is $\sin$. At the end of the 5000 epochs [see Fig.~\ref{fig-LDEa}], the recorded mean squared error (MSE) loss was on the order of \(10^{-6}\). Further details are listed in Table~\ref{TabelLinearODE}. The comparison of the PIKAN Predicted solution and the exact Numerical result of Eq~\eqref{EqODE11} are given in Fig.~\ref{fig-LDEb}.

\begin{table}
\centering
\normalsize
\resizebox{\linewidth}{!}{
\begin{tabular}{c|c|c|c|c}
  \hline
  \textbf{Architecture} & \textbf{Spline-order} & \textbf{Grid size} & \textbf{Basis Activation} & \textbf{Loss} \\
  \hline
  \hline
  $[1, 5, 4, 3, 1]$ & 3 & 5 & $\sin$ & $10^{-6}$ \\
  \hline
\end{tabular}}
\caption{The architecture details for Sec.~\ref{SecODESingle}.}\label{TabelLinearODE}
\end{table}

\section{System of Ordinary Differential Equations}\label{COUPLED_DEQN}

In this section, we employed PIKAN to solve coupled linear and non-linear differential equations that have been previously solved using PINN \cite{cde1} and by Galerkin Method with Cubic B-Splines \cite{cde2}. 

We begin by considering the simple case 
\begin{equation}
\begin{aligned}
\frac{dv(x)}{dx} &= u, \\
\frac{du(x)}{dx} &= -v,
\label{coupled_simple}
\end{aligned}
\qquad
\left\{
\begin{aligned}
u(0) &= 1, \\
v(0) &= 0.
\end{aligned}
\right.
\end{equation}
For which the residuals are defined as
\begin{equation}
\mathcal{R}^{u}_\theta(t) = \frac{d\hat{u}(x)}{dx} + \hat{v}(x) , \quad  \mathcal{R}^{v}_\theta(t) = \frac{d\hat{v}(x)}{dx} - \hat{u}(x). \nonumber
\end{equation}
Note that $\hat{u}(x)$ and $\hat{v}(x)$ are the PIKAN solution to be learned. Note that henceforth, we drop  the parameter $\theta$ in the notation of the PIKAN solution.  equations were solved using PINN \cite{cde1} and by Galerkin Method with Cubic B-Splines \cite{cde2}. Hereinafter, the weights $\lambda_{\mathrm{r}}, \lambda_{\mathrm{ic}}$ and $\lambda_{\mathrm{bc}}$ in the final loss function~\eqref{weight-loss-DD} are all set to unity. As a result, the final loss becomes
\begin{align}
  \mathcal{L}^{DF}_{\text{final}} = \frac{1}{N_{r}} \sum_{i=1}^{N_{r}} \left( \left| \mathcal{R}^{u}_\theta(x_{\mathrm{r}}^{i}) \right| ^2 + \left| \mathcal{R}^{v}_\theta(x_{\mathrm{r}}^{i}) \right| ^2 \right) \nonumber \\  + 
  (\hat{u}(0) - 1)^2 + (\hat{v}(0) - 0)^2.  
  \label{LOSS_CDE}
\end{align}

\begin{figure}[h!]
    \centering
    \begin{subfigure}[t]{0.5\textwidth}
        \centering
        \includegraphics[width=\textwidth]{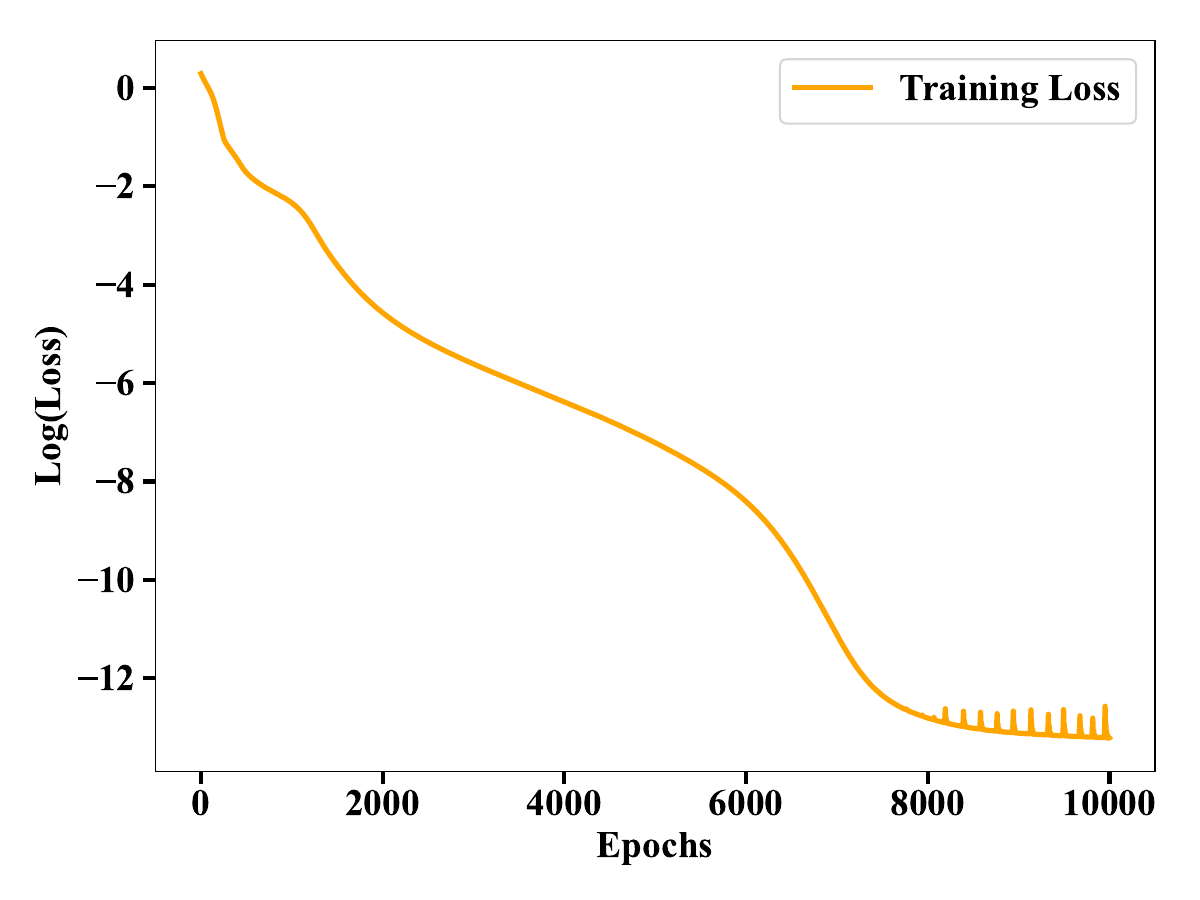}
        \caption{}
        \label{fig-CDEa}
    \end{subfigure}%
    \hfill
    \begin{subfigure}[t]{0.5\textwidth}
        \centering
        \includegraphics[width=\textwidth]{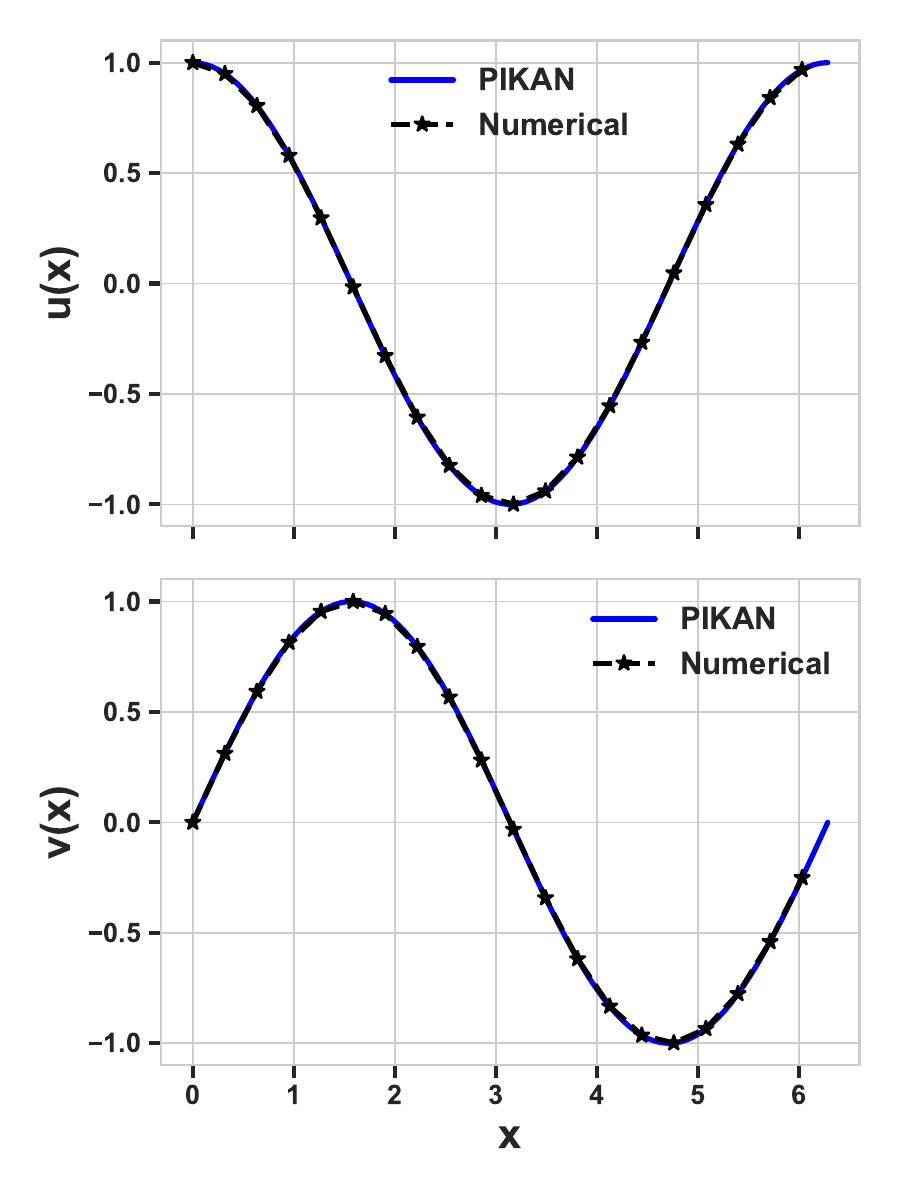}
        \caption{}
        \label{fig-CDEb}
    \end{subfigure}
    \caption{(a) Illustrates the relationship between loss~\eqref{LOSS_CDE} and epoch, (b) A comparison between PIKAN predicted solution and numerically exact solution of Eq.~\eqref{coupled_simple}.}
\end{figure}

\begin{table}
\centering
\normalsize
\begin{tabular}{c|c|c}
  \hline
  \textbf{Architecture}  & \textbf{Wavelet type} & \textbf{Loss} \\ 
  \hline
  \hline
  $[1,3,2]$  & $\sin$ & $10^{-6}$ \\ 
  \hline
\end{tabular}
\caption{Architecture details of the model used in section~\ref{COUPLED_DEQN}.}
\label{CDE_TABLE_01}.
\end{table}

In coupled differential equation we have seen that the convergence of the efficient-KAN is not satisfactory and to resolve this, we have used WAV-KAN employing  collocation points $N_{x} = 100$ equally distributed on the interval $0 \leq x \leq  2\pi$. The model was trained for 10000 epochs and Adam optimizer with a learning-rate $\eta = 0.001$ is used.The recorded mean squared error (MSE)
loss was of the order of $10^{-6}$. Further details can be obtained from Table~\ref{CDE_TABLE_01}.
The comparison of the PIKAN predicted  solution and the numerically calculated  solution of Eq.~\eqref{coupled_simple} is shown in Fig~\ref{fig-CDEb}.

\subsection{System of Linear Ordinary Differential Equations}\label{Linear_Coupled_single_Variable}

Consider the linear coupled system with boundary conditions \cite{cde1}:
\begin{align}
    \begin{aligned}\label{LCDE_17}
        u''(x) + xu(x) + 2v'(x) &= u_1, \\
        u(x) + v''(x) + 2v(x) &= u_2,  \nonumber
    \end{aligned} \\
    \begin{cases}
        u(0) &= 0 \\
        v'(1) + v(1) &= \cos(1) + \sin(1) \\
        u(1) &= 2 \\
        v'(0) &= 1. 
    \end{cases}
\end{align}
Here, $0 \leq x \leq 1$ and the non-homogeneous terms are $ u_1(x) = x^3 + x^2 + 2 + 2\cos(x)$  and  $ u_2(x) = x^2 + x + \sin(x) $,
 respectively.
 
The residuals read
\begin{equation*}
\begin{aligned}
\mathcal{R}^{1}_\theta(x) &=  \hat{u}''(x) + x \hat{u}(x) + 2 \hat{v}'(x) - u_{1}(x) , \\
\mathcal{R}^{2}_\theta(x) &=  \hat{u}(x) + \hat{v}''(x) + 2 \hat{v}(x) - u_{2}(x) .
\end{aligned}
\end{equation*}
As a result, we have the physics loss, the boundary loss and the final loss as 
\begin{equation}
\begin{aligned}
    \mathcal{L}_{\mathrm{r}} &=  \frac{1}{N_{r}} \sum_{i=1}^{N_{r}} \left( \left| \mathcal{R}^{1}_\theta(x_{\mathrm{r}}^{i}) \right| ^2 + \left| \mathcal{R}^{2}_\theta(x_{\mathrm{r}}^{i}) \right| ^2 \right), \\
    \mathcal{L}_{\mathrm{bc}} &= \left( \hat{u}(0) - 0 \right)^2 + \left( \hat{v}'(1) + \hat{v}(1) - (\cos(1) + \sin(1)) \right)^2 \\
    &\quad + \left( \hat{u}(1) - 2 \right)^2 + \left( \hat{v}'(0) - 1 \right)^2, \\
    \mathcal{L}^{DF}_{\mathrm{final}} &= \mathcal{L}_{\mathrm{r}} + \mathcal{L}_{\mathrm{bc}}\label{LCDE_LOSS}.
\end{aligned}
\end{equation}

\begin{figure}[h!]
    \centering
    \begin{subfigure}[t]{0.48\textwidth}
        \centering
        \includegraphics[width=\textwidth]{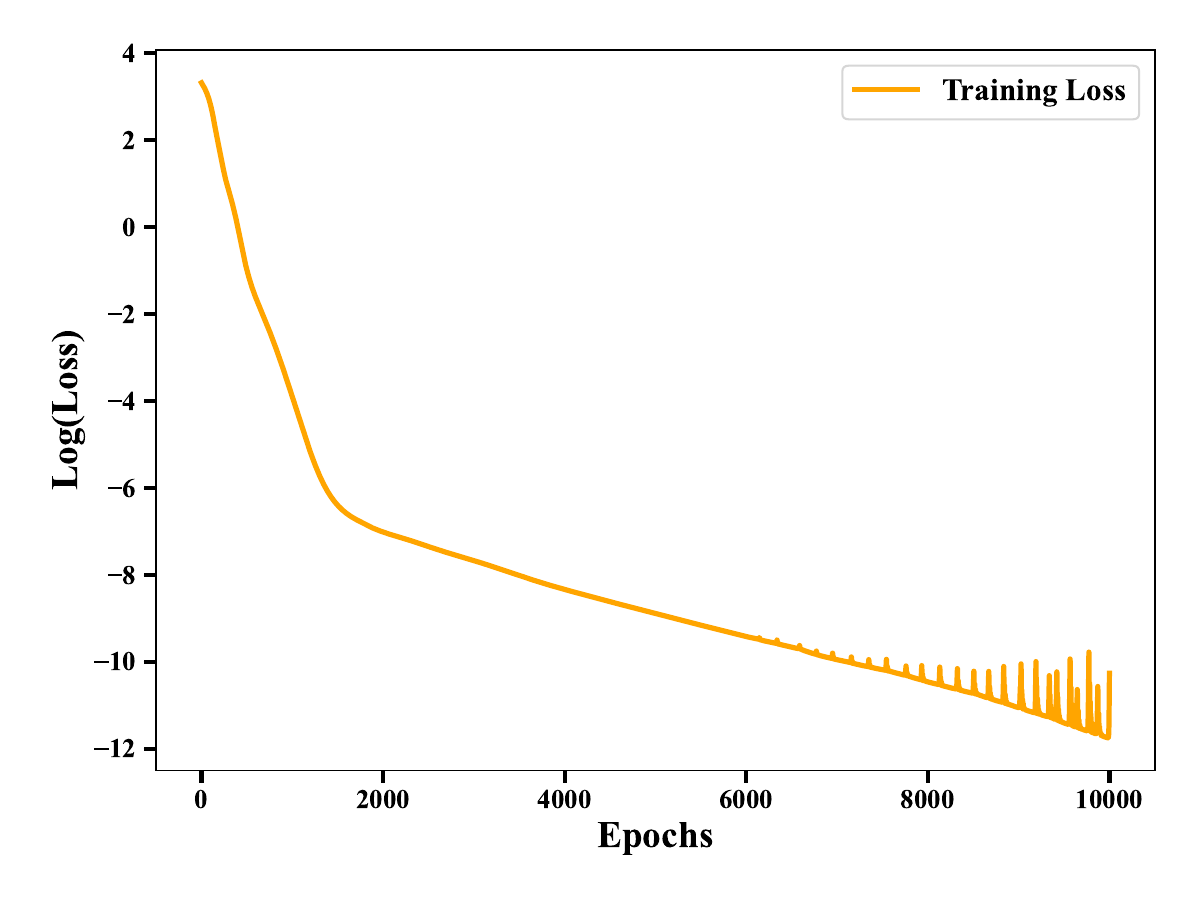}
        \caption{}
        \label{fig-CDE1a}
    \end{subfigure}%
    \hfill
    \begin{subfigure}[t]{0.48\textwidth}
        \centering
        \includegraphics[width=\textwidth]{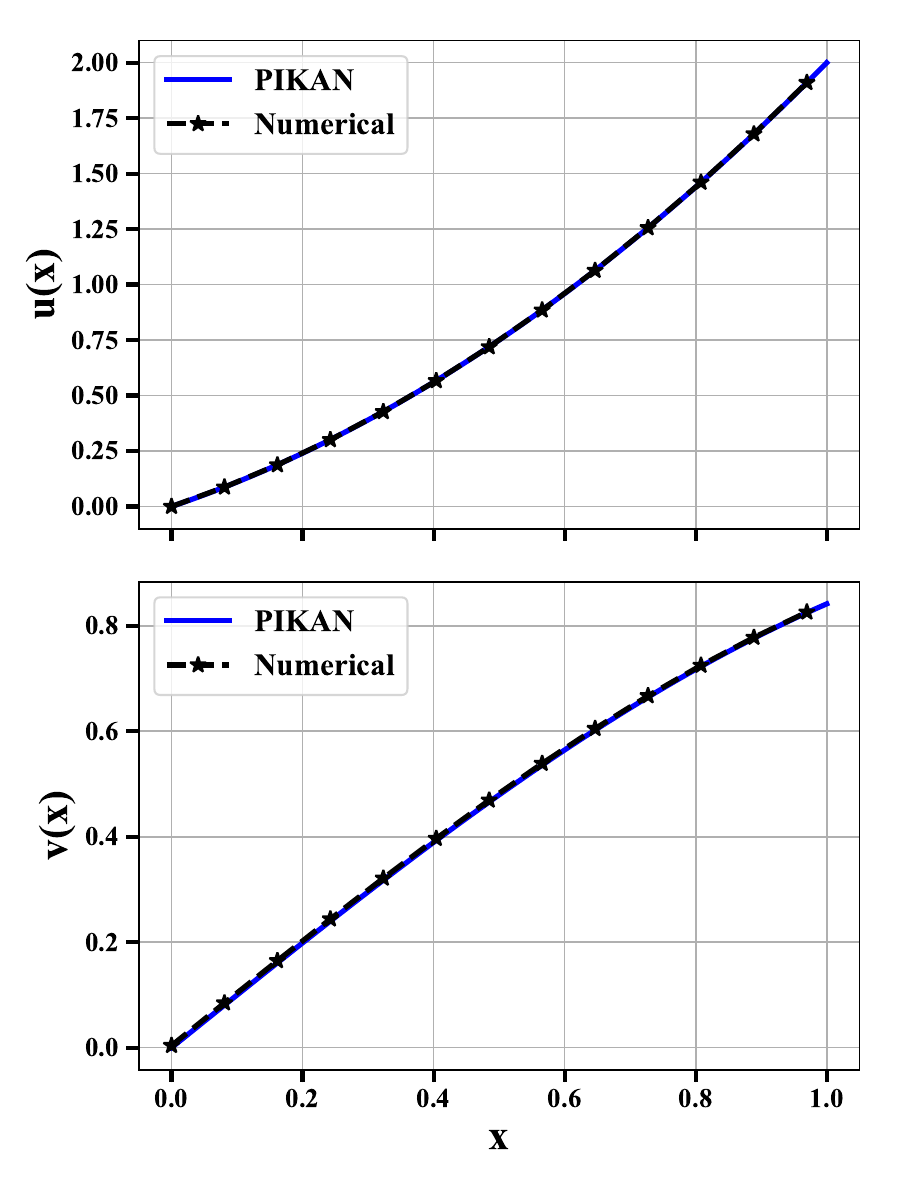}
        \caption{}
        \label{fig-CDE1b}
    \end{subfigure}
    \caption{(a) Illustrates the relationship between loss~\eqref{LCDE_LOSS} and epoch, (b) A comparison between the PIKAN predicted solution and numerically exact solution of Eq.~\eqref{LCDE_17}.}
\end{figure}

We have utilized the efficient-KAN method to solve the coupled-linear differential equation~\eqref{LCDE_17}, employing collocation points $N_{x} = 100$ equally distributed on the interval $0\leq x \leq 1$
\cite{cde1}. Note that we have observed that there is not much difference between the performance of WAV-KAN and efficient-KAN for this problem, but the convergence of WAV-KAN is faster than efficient-KAN.
The model was trained for 10000 epochs
and  AdamW optimizer is used  with a learning rate $\eta= 0.001$. At the end of the 10000 epochs, the recorded loss was of the order of $10^{-5}.$
Further details about the model used are listed in Table~\ref{LCDE_Table}. The comparison of the PIKAN predicted solution and the numerically exact solution of $u(x)$ and $v(x)$ is shown in Fig~\ref{fig-CDE1b}.
\begin{table}
\centering
% \normalsize
\begin{tabular}{c|c|c|c|c}
  \hline
  Architecture & Spline-order & Grid size & Basis Activation & Loss \\
  \hline
  \hline
  $[1, 2,3,2]$ & 3 & 5 & $\sin$ & $10^{-5}$ \\
  \hline
\end{tabular}
\caption{Architecture details of the model used in section~\ref{Linear_Coupled_single_Variable}.}
\label{LCDE_Table}
\end{table}

\subsection{System of Nonlinear Ordinary Differential Equations}\label{Nonlinear_Coupled_diffferentialequation_sollution_technique}

Consider the problem\cite{cde1}:

\begin{equation}
\begin{aligned}\label{Nonlinear_Deqn}
u''(x) + xu(x) + 2xv(x) + xu^2(x) &= u_3(x), \\
x^2 u(x) + v'(x) + v(x) + \sin(x) v^2(x) &= u_4(x), \\
\begin{cases}
u(0) = 0, \quad u(1) = 0, \\ 
v(0) = 0, \quad v(1) = 0.
\end{cases}
\end{aligned}
\end{equation}
where, 
\begin{align}\label{Nonlinear_Coupled}
    u_3(x) &= 2x \sin(\pi x) + x^5 - 2x^4 + x^2 - 2, \nonumber \\
    u_4(x) &= x^3 (1 - x) + \sin(\pi x) \left(1 + \sin(x) \sin(\pi x)\right) \nonumber \\
           &\quad + \pi \cos(\pi x). \nonumber
\end{align}
Let us define the residual $ \mathcal{R}^{1}_\theta(x)$  and $ \mathcal{R}^{2}_\theta(x)$ using PIKAN to-be-learned solutions $\hat{u}(x)$ and $\hat{v}(x)$ as
\begin{equation*}
\begin{aligned}
\mathcal{R}^{1}_\theta(x) &=  \hat{u}''(x) + x \hat{u}(x) + 2x \hat{v}'(x) - u_{3}(x), \\
\mathcal{R}^{2}_\theta(x) &=  \hat{u}(x) + \hat{v}''(x) + 2\hat{v}(x) - u_{4}(x).
\end{aligned}
\end{equation*}

As a result, the physics loss, boundary loss and final loss becomes
\begin{align}
\mathcal{L}_{\mathrm{r}} &= \frac{1}{N_{r}} \sum_{i=1}^{N_{r}} \left( \left| \mathcal{R}^{1}_\theta(x_{\mathrm{r}}^{i}) \right| ^2 + \left| \mathcal{R}^{2}_\theta(x_{\mathrm{r}}^{i}) \right| ^2 \right), \\
    \mathcal{L}_{bc} = &\left( \hat{u}(0) - 0 \right)^2 + \left( \hat{u}(1) - 0 \right)^2 \nonumber\\ 
    &+ \left( \hat{v}(0) - 0 \right)^2 + \left( \hat{v}(1) - 0 \right)^2,  \\
 \mathcal{L}^{DF}_{\mathrm{final}} &= \mathcal{L}_{\mathrm{r}} + \mathcal{L}_{\mathrm{bc}}.\label{NonlinearODE_LOSS}
\end{align}

\begin{figure}
    \centering
    \begin{subfigure}[t]{0.5\textwidth}
        \centering
        \includegraphics[width=\textwidth]{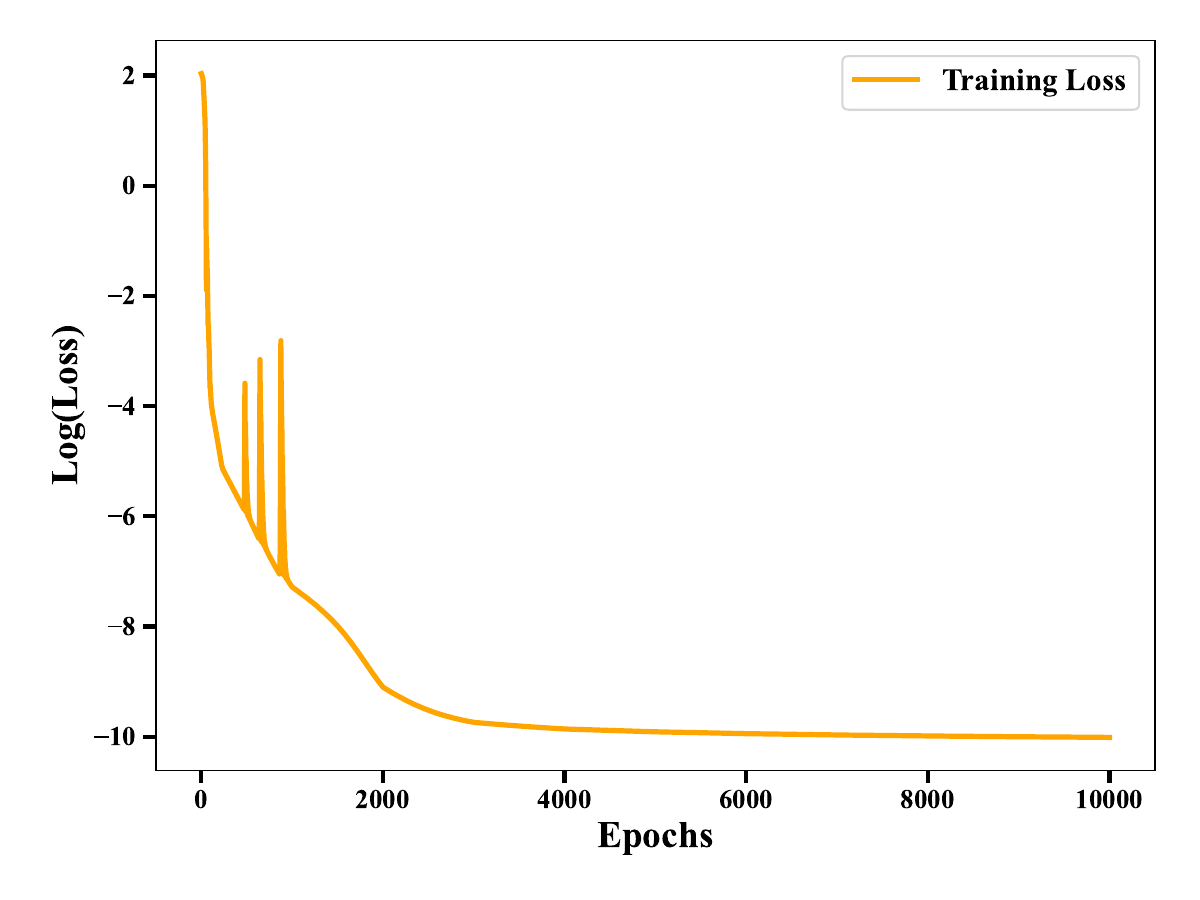}
        \caption{}
        \label{fig-NCDEa}
    \end{subfigure}%
    \hfill
    \begin{subfigure}[t]{0.48\textwidth}
        \centering
        \includegraphics[width=\textwidth]{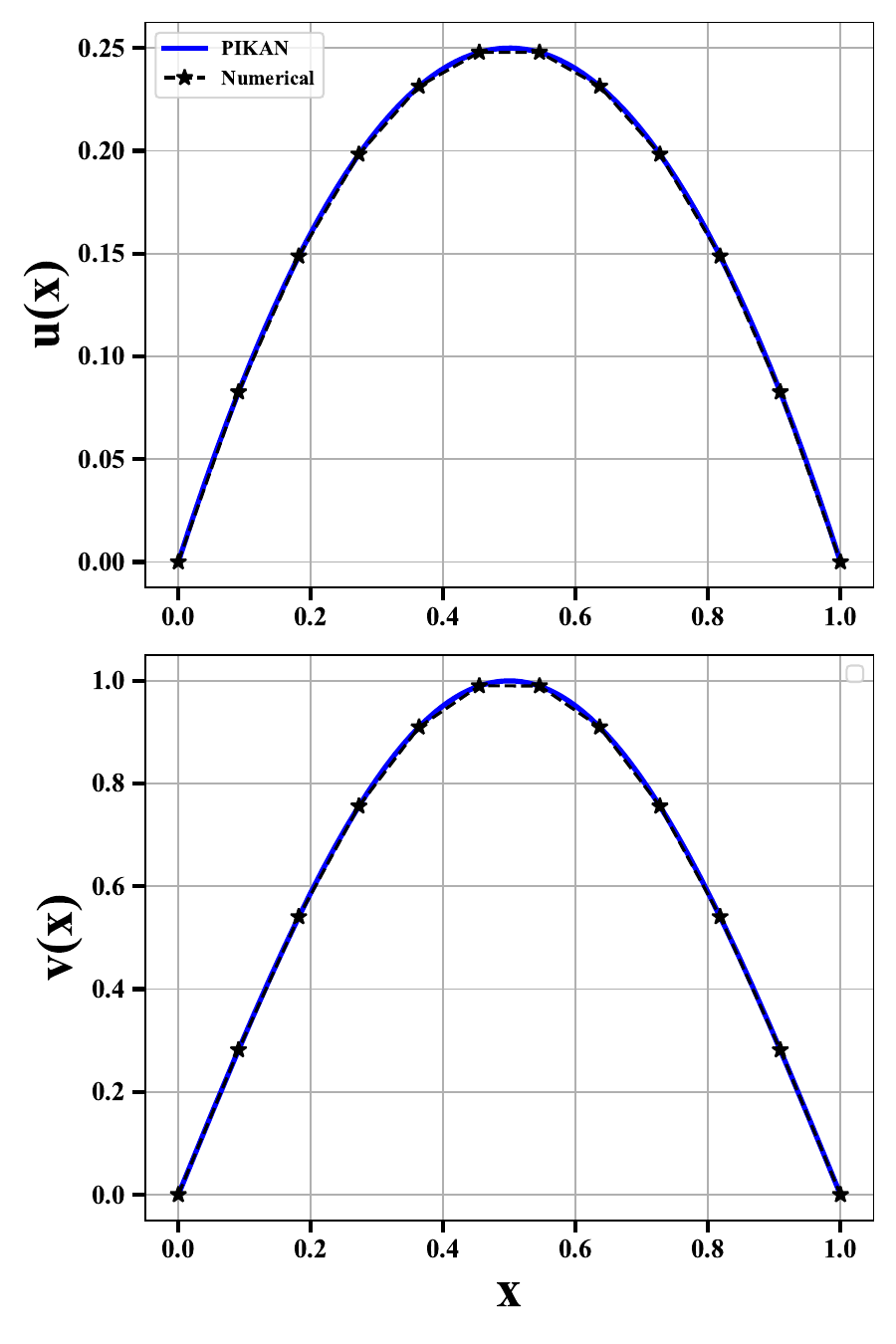}
        \caption{}
        \label{fig-NCDEb}
    \end{subfigure}
    \caption{(a) Illustrates the relationship between loss~\eqref{NonlinearODE_LOSS} and epoch, (b) A comparison between PIKAN predicted solution and numerically exact solution of Eq~\eqref{Nonlinear_Deqn}.}
    \label{Nonlinear_Coupled_ODE}
\end{figure}

We have utilized the WAV-KAN method to solve the coupled-nonlinear differential equation~\eqref{Nonlinear_Deqn}, employing collocation points $N_{x} = 100$ equally distributed on the interval $0\leq x \leq 1$
\cite{cde1}. The interesting point is that there is not much difference between the performance of WAV-KAN and efficient-KAN for this particular problem but the convergence rate of WAV-KAN was faster than efficient-KAN. The model was trained for 10000 epochs and  AdamW optimizer is used  with a variable learning rate starting from  $\eta= 0.01$, and at the end of every 1000 epochs the learning rate is updated following the given rule, $\eta_{new} = \frac{\eta_{old}}{5}$ . At the end of the 10000 epochs, the recorded loss was of the order of $10^{-5}.$
Further details about the model used is listed in Table~\ref{Nonlinear_coupled_Table}. The comparison of PIKAN predicted solution and the numerically computed solution is shown in Fig~\ref{fig-NCDEb}.

\begin{table}[htbp]
\centering
\begin{tabular}{c|c|c}
  \hline
  Architecture  & Wavelet type & Loss \\
  \hline
  \hline
  $[1,7,2]$  & $\sin$ & $10^{-5}$ \\
  \hline
\end{tabular}
\caption{Architecture details of the model used in section~\ref{Nonlinear_Coupled_diffferentialequation_sollution_technique}.}
\label{Nonlinear_coupled_Table}
\end{table}

\subsection{The Lorentz Equations}\label{Lorentz_System}

The Lorenz equation is a system of ordinary differential equations that was originally developed to model convection currents in the atmosphere \cite{Lorenz2}. It is an iconic example in the chaos theory and has significantly influenced our understanding of dynamic systems. We focus on exploring the non-chaotic solutions of the Lorentz system. Consider the following set of equations \cite{Lorenz1},
\begin{equation}
    \begin{aligned}
        x'(t) &= \sigma (y(t) - x(t)), \\
        y'(t) &= x (\rho - z(t)) - y(t), \\
        z'(t) &= x(t)y(t) - \beta z(t).
    \end{aligned}
    \qquad
    \begin{cases}
        x(0) = 1, \\
        y(0) = 1, \\
        z(0) = 1,
    \end{cases}
    \quad
    \label{lorentz_equation}
\end{equation}
Let us define the following residuals using to-be-learned PIKAN solutions $\hat{x}(t)$, $\hat{y}(t)$ and $\hat{z}(t)$ as,
\begin{equation}
    \begin{aligned}
    \mathcal{R}^{1}_\theta(t) &=   \hat{x}'(t) - \sigma (\hat{y}(t) - \hat{x}(t)) \\
    \mathcal{R}^{2}_\theta(t) &=  \hat{y}'(t) - ( \hat{x}(t) (\rho - \hat{z}(t)) -\hat{y}(t) )
    \\
    \mathcal{R}^{3}_\theta(t) &=  \hat{z}'(t) - (\hat{x}(t) \hat{y}(t) - \beta \hat{z}(t)) 
    \end{aligned}
\end{equation}
and the physics loss, initial loss and final loss function takes the form:
\begin{equation}
\begin{aligned}\label{Lorentz_loss}
\mathcal{L}_{\mathrm{r}} &= \frac{1}{N_{r}} \sum_{i=1}^{N_{r}} \left( \left| \mathcal{R}^{1}_\theta(t_{\mathrm{r}}^{i}) \right|^2 + \left| \mathcal{R}^{2}_\theta(t_{\mathrm{r}}^{i}) \right|^2 + \left| \mathcal{R}^{3}_\theta(t_{\mathrm{r}}^{i}) \right|^2 \right), \\
\mathcal{L}_{ic} &= (\hat{x}(0) - 1)^2 + (\hat{y}(0) - 1)^2 + (\hat{z}(0) - 1)^2, \\
\mathcal{L}^{DF}_{\mathrm{final}} &= \mathcal{L}_{\mathrm{r}} + \mathcal{L}_{\mathrm{ic}}.
\end{aligned}
\end{equation}
\begin{figure}[h!]
    \centering
    \begin{subfigure}[t]{0.5\textwidth}
        \centering
        \includegraphics[width=\textwidth]{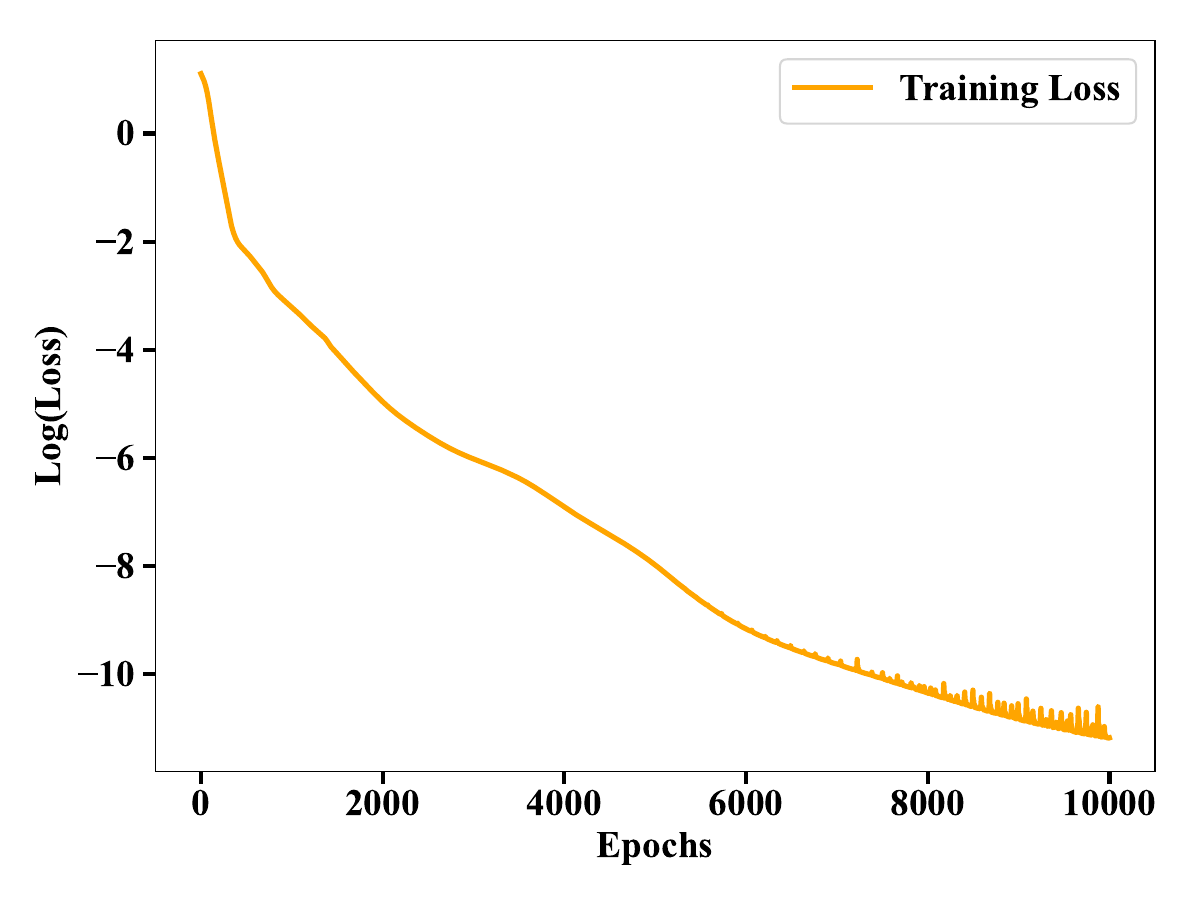}
        \caption{}
        \label{fig-lorentz_1a}
    \end{subfigure}%
    \hfill
    \begin{subfigure}[t]{0.5\textwidth}
        \centering
        \includegraphics[width=\textwidth]{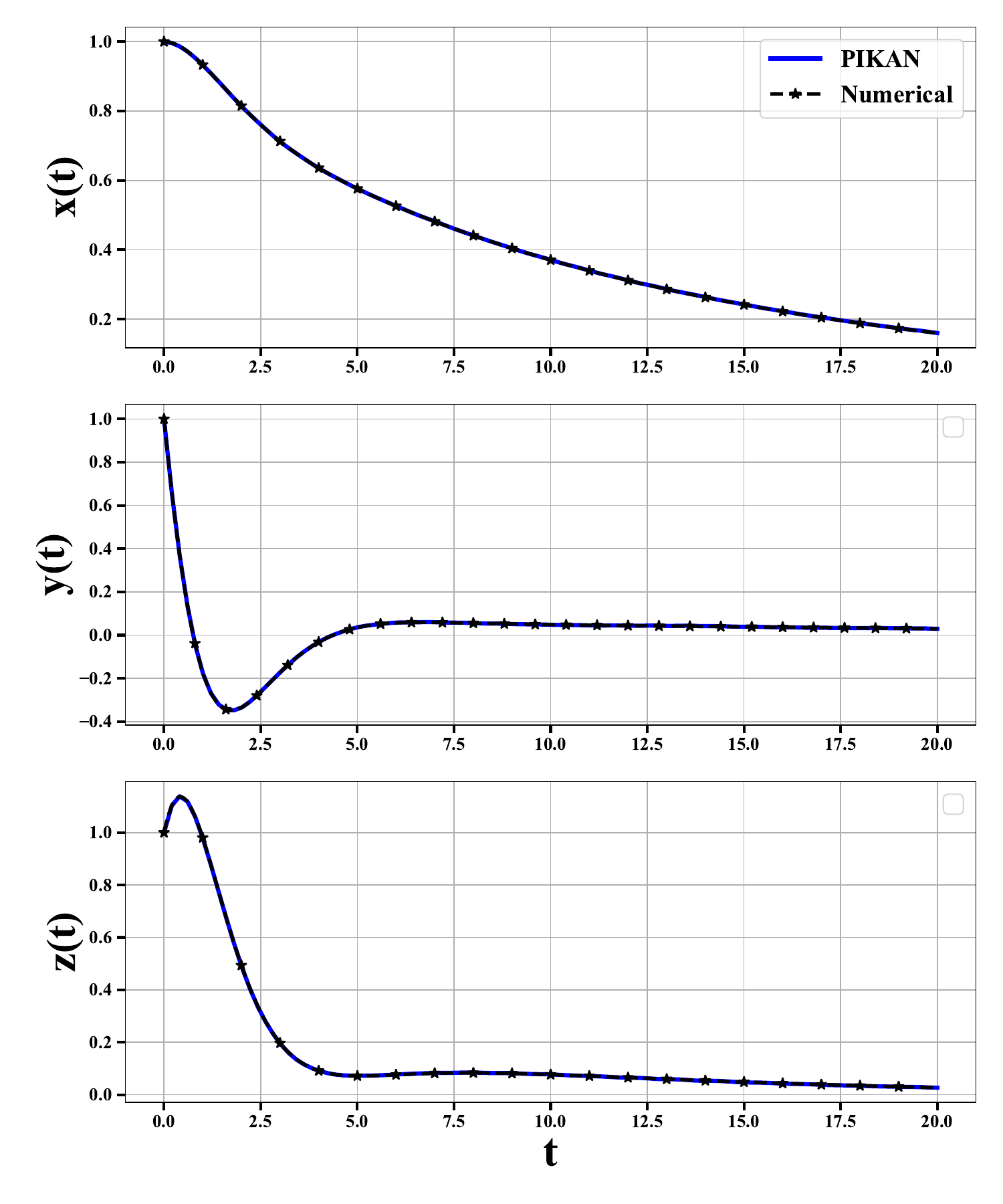}
        \caption{}
        \label{fig-lorentz_1b}
    \end{subfigure}
    \caption{(a) Illustrates the relationship between loss~\eqref{Lorentz_loss} and epoch.(b) Comparison between the PIKAN predicted solution and numerically exact solution of  Eq.~\eqref{lorentz_equation}.}
\end{figure}
To solve this differential equation we have used both WAV-KAN and efficient-KAN, however the performance of WAV-KAN was better than the efficient-KAN. We have taken collocation points $N_r = 100$ equally distributed over the interval $0\leq t \leq 20$. The model was trained for $10^4$ epochs and to minimize the loss function we have used used the AdamW optimizer with a learning rate $\eta = 0.001$. At the end of learning [Fig.~\ref{fig-lorentz_1a}], the recorded mean squared loss (MSE) was on the order of $10^{-5}$. Further details of the model can be obtained from the Table~\ref{Lorentz_Table}. The comparison between the PIKAN predicted solution and the numerical solutions are shown in Fig.~\ref{fig-lorentz_1b}.

\begin{table}
\centering
\begin{tabular}{c|c|c}
  \hline
  Architecture  & Wavelet type & Loss \\
  \hline
  \hline
  $[1, 8,16,3]$  & $\sin$ & $10^{-5}$ \\
  \hline
\end{tabular}
\caption{Architecture details of the model used in section~\ref{Lorentz_System}.}
\label{Lorentz_Table}
\end{table}

\section{Oscillatory dynamics} \label{osc_dyn}

\subsection{Simple Harmonic Oscillator}\label{SHM}

Consider the problem
\begin{equation}
    y''(t) + \omega_0^2y(t) = 0, \quad
    \begin{cases}
    y(0) = 0.1, \\
    y'(0) = 40,
    \end{cases}
    \label{SHM_DEQN}
\end{equation}
where $ \omega_0 = 25 $ is the normalized angular frequency , where $0 \leq  t \leq 1$. Let us define the following residual
\begin{equation*}
\begin{aligned}
    \mathcal{R}_\theta(t)&=  \hat{y}''(t) +  \omega_{0}^2 \hat{y}(t)
    \end{aligned}
\end{equation*}
 Consequently, the physics-loss term, the initial condition loss term and the final loss terms are 
\begin{align}
\mathcal{L}_{\mathrm{r}} &= \frac{1}{N_{r}} \sum_{i=1}^{N_{r}} \left|\mathcal{R}_\theta(t_{\mathrm{r}}^{i})\right |^2, \nonumber \\ \nonumber
\mathcal{L}_{\mathrm{ic}} &= \left[(\hat{y}(0) - 0.1)\right)^2 + \left( \hat{y}'(0) - 40 \right)^2] 
\\
\quad \mathcal{L}_{\mathrm{final}}^{DF} &= \mathcal{L}_{\mathrm{r}} + \mathcal{L}_{\mathrm{ic}}
\label{SHM_LOSS}
\end{align}
\begin{figure}[h!]
    \centering
    \begin{subfigure}[t]{0.5\textwidth}
        \centering
        \includegraphics[width=\textwidth]{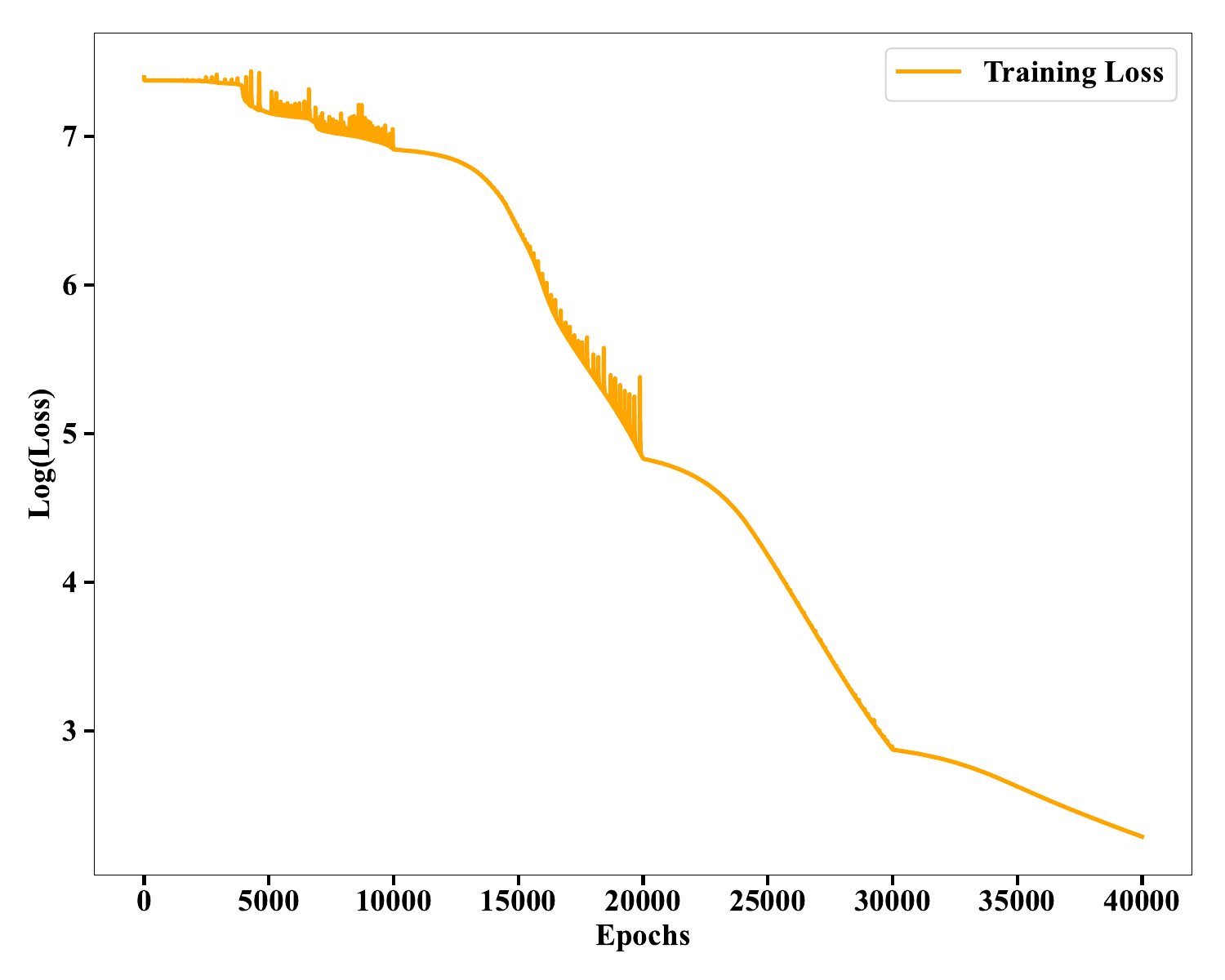}
        \caption{}
        \label{fig-SHM_1a}
    \end{subfigure}%
    \hfill
    \begin{subfigure}[t]{0.5\textwidth}
        \centering
        \includegraphics[width=\textwidth]{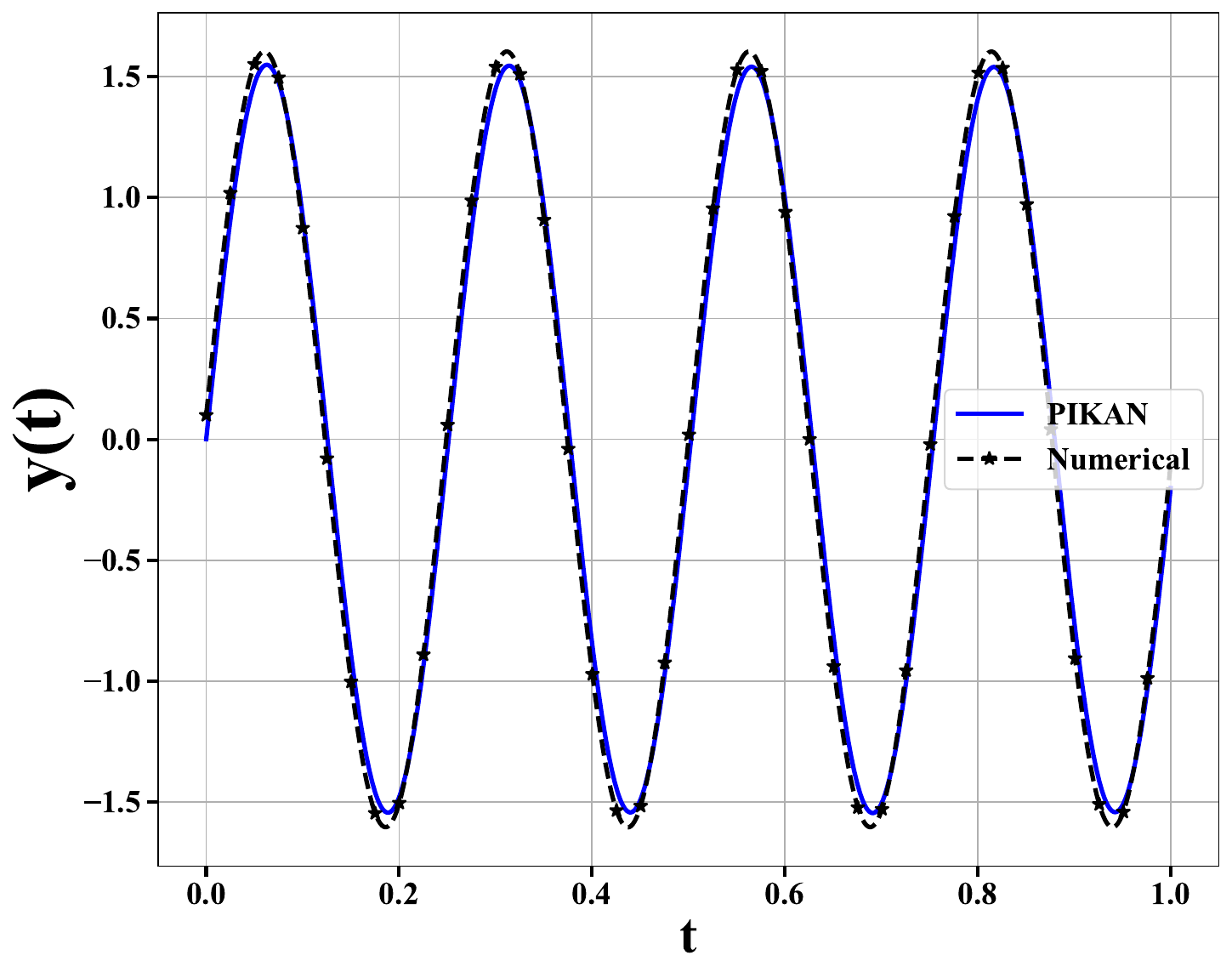}
        \caption{}
        \label{fig-SHM_1b}
    \end{subfigure}
    \caption{(a) Illustrates the relationship between loss~\eqref{SHM_LOSS} and epoch.(b) Comparison between the PIKAN predicted solution and numerically exact solution for  Eq.~\eqref{SHM_DEQN}.}
\end{figure}

We have utilized WAV-KAN to solve the Eq~\eqref{SHM_DEQN} employing the collocation point $N_t = 100$ equally distributed over the interval $0 \leq t \leq 1$. We have trained the neural network for 40000 epochs and we have used Adam optimizer with a variable learning rate $\eta$ starting from $0.001$ and  decreasing $\eta$ by 10 after every $10^4$ epochs. At the end of 40000 epochs, the recorded mean squared error (MSE) loss was on the order of $10^{-1}$. Further details of the model can be obtained from Table~\ref{SHM_table}. The comparative analysis of the PIKAN predicted solution and the numerically exact solution are given in Fig.~\ref{fig-SHM_1b}.
\begin{table}
\centering
\begin{tabular}{c|c|c}
  \hline
  Architecture  & Wavelet type & Loss \\
  \hline
  \hline
  $[1,8,6,8,1]$  & $\sin$ & $0.2$ \\
  \hline
\end{tabular}
\caption{Architecture details of the model used for the section~\ref{SHM}. }
\label{SHM_table}
\end{table}

\subsection{Nonlinear Pendulum}\label{Nonlinear_Shm}

The nonlinear pendulum Eq.~\eqref{Nonlinear_Dqn} with a sine function, provides an exact depiction of the pendulum's dynamics for all oscillation amplitudes;

\begin{equation}
    y''(t) + \omega_0^2\sin(y(t)) = 0, \quad
    \begin{cases}
    y(0) = 0.1, \\
    y'(0) = 40,
    \end{cases}
    \label{Nonlinear_Dqn}
\end{equation}
where $ \omega_0 = 25 $ is the normalized angular frequency, where $0 \leq t \leq 1$. 

Let us define the residual as
\begin{equation*}
\begin{aligned}
    \mathcal{R}_\theta(t)&=  \hat{y}''(t) +  \omega_{0}^2 \sin(\hat{y}(t))
    \end{aligned}
\end{equation*}
and the physics-loss, the initial condition loss as well as the the final loss are given by
\begin{align}
\mathcal{L}_{\mathrm{r}} &= \frac{1}{N_{r}} \sum_{i=1}^{N_{r}} \left|\mathcal{R}_\theta(t_{\mathrm{r}}^{i})\right |^2, \nonumber \\ \nonumber
\mathcal{L}_{\mathrm{ic}} &= \left[(\hat{y}(0) - 0.1)\right)^2 + \left( \hat{y}'(0) - 40 \right)^2], 
\\
\quad \mathcal{L}_{\mathrm{final}}^{DF} &= \mathcal{L}_{\mathrm{r}} + \mathcal{L}_{\mathrm{ic}}.
\label{Nonlinear_loss}
\end{align}
\begin{figure}[h!]
    \centering
    \begin{subfigure}[t]{0.5\textwidth}
        \centering
        \includegraphics[width=\textwidth]{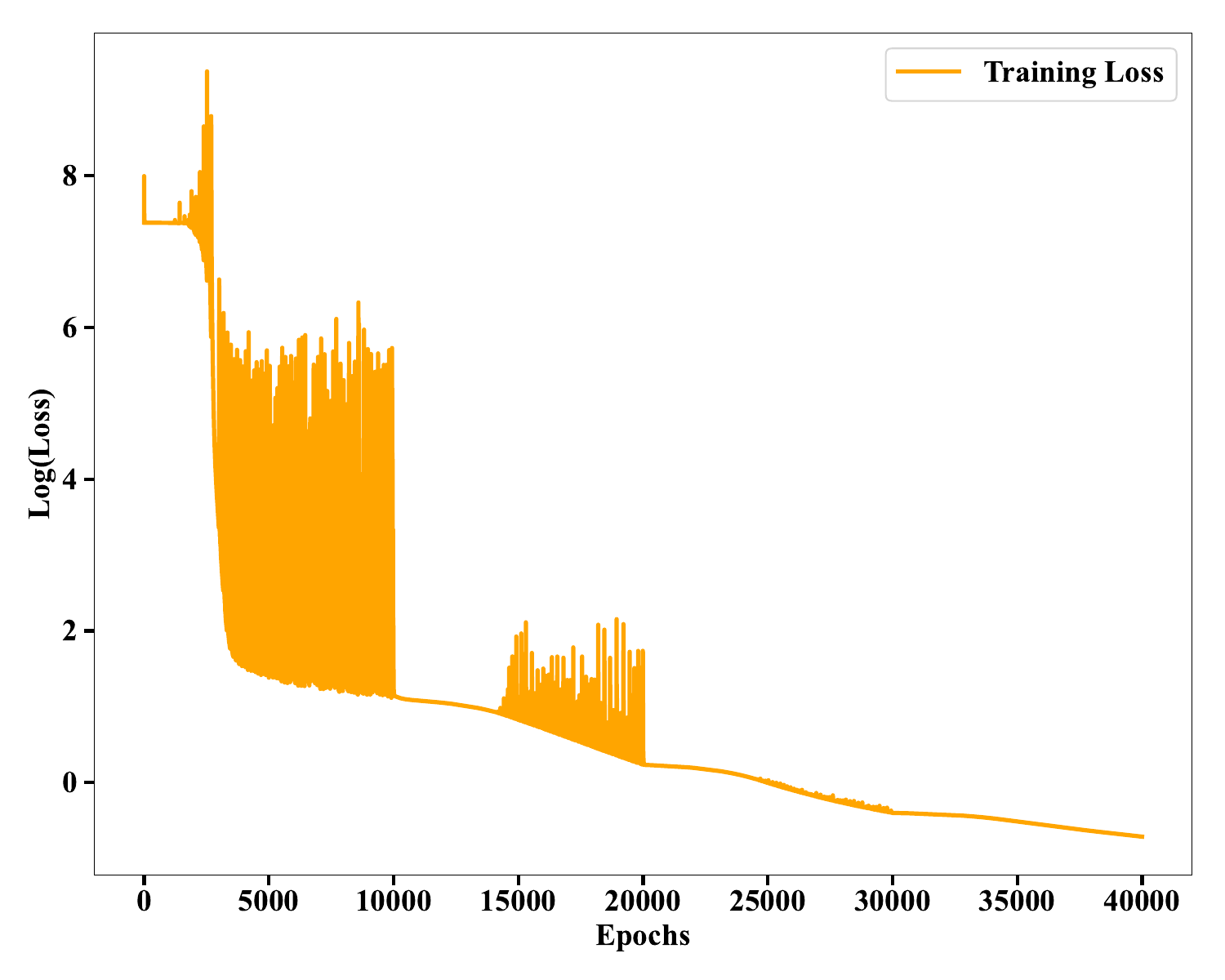}
        \caption{}
        \label{fig-NSHM_1a}
    \end{subfigure}%
    \hfill
    \begin{subfigure}[t]{0.5\textwidth}
        \centering
        \includegraphics[width=\textwidth]{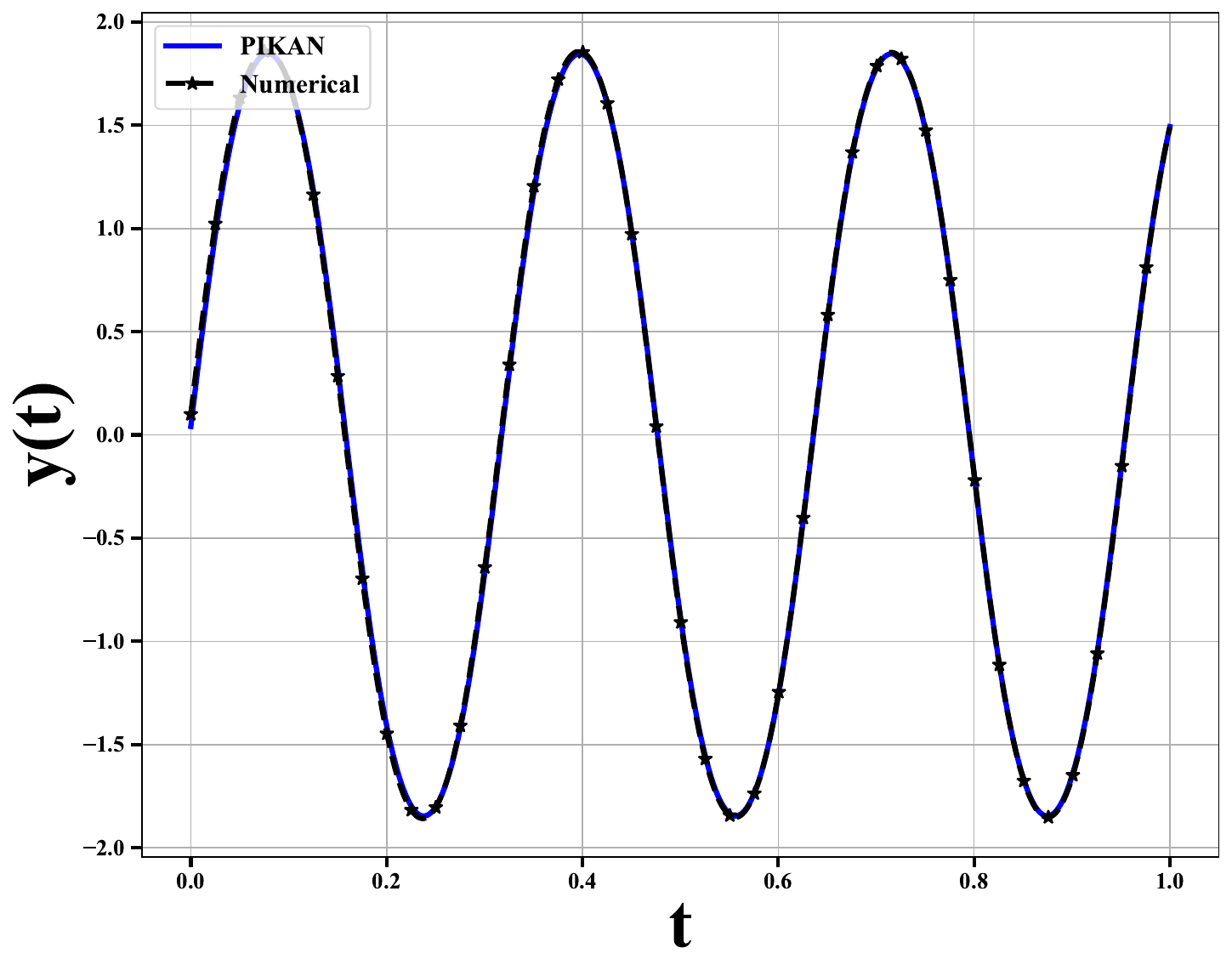}
        \caption{}
        \label{fig-NSHM_1b}
    \end{subfigure}
    \caption{(a) Illustrates the relationship between loss~\eqref{Nonlinear_loss} and epoch.(b) Comparison between the PIKAN predicted solution and numerically exact solution of  Eq.~\eqref{Nonlinear_Dqn}.}
\end{figure}
We have utilized WAV-KAN to solve the Eq.~\eqref{Nonlinear_Dqn} employing the collocation point $N_t = 100$ equally distributed over the interval $0 \leq t \leq 1$.
We have trained the neural network for 40000 epochs and we have used Adam optimizer with a variable learning rate $\eta$ starting from $0.001$ and  decreasing $\eta$ by 10 after every $10^4$ epochs. At the end of 40000 epochs, the recorded mean squared error loss was on the order of $10^{-1}$. Further details of the model can be obtained from Table~\ref{NSHM_table}. The comparative analysis of the PIKAN predicted solution and the numerically exact solution are given in Fig~\ref{fig-NSHM_1b}.
 \begin{table}
\centering
\begin{tabular}{c|c|c}
  \hline
  Architecture  & Wavelet type & Loss \\
  \hline
  \hline
  $[1,8,6,8,1]$  & $\sin$ & $0.9$ \\
  \hline
\end{tabular}
\caption{Architecture details of the model used in section~\ref{Nonlinear_Shm}.}
\label{NSHM_table}
\end{table}

\subsection{Mathieu Equation}
\label{Mathew_eqn}
Mathieu's equation stems from his 1868 study on vibrations in an elliptical drum \cite{Mathieu0}. Mathieu’s equation is a linear second-order ordinary differential equation distinguished from a simple harmonic oscillator by its time-varying (periodic) stiffness coefficient \cite{Mathieu1, Mathieu2}.
Consider the initial value problem \cite{Rahman2024}:
\begin{equation}
    \label{Mathew_deqn}
    y''(t) + (a + \beta \cos t) y(t) = 0, \quad
    \begin{cases}
    y(0) = 1, \\
    y'(0) = 0,
    \end{cases}
\end{equation}
The residual corresponding to this equation is
\begin{equation*}
    \begin{aligned}
    \mathcal{R}_\theta(t)&=  \hat{y}''(t) + (a + \beta \cos t) \hat{y}(t)
    \end{aligned}
\end{equation*}
Consequently the physics-loss term, the initial condition loss term, and the total-loss terms are 
\begin{equation}
    \begin{aligned}
    \mathcal{L}_{\mathrm{r}} &= \frac{1}{N_{r}} \sum_{i=1}^{N_{r}} \left|\mathcal{R}_\theta(t_{\mathrm{r}}^{i})\right |^2, \\
    \mathcal{L}_{\mathrm{ic}} &= \left[(\hat{y}(0) - 1)\right)^2 + \left( \hat{y}'(0) \right)^2]
    \\ 
\mathcal{L}_{\mathrm{final}}^{DF} &= \mathcal{L}_{\mathrm{r}} + \mathcal{L}_{\mathrm{ic}}
    \label{Mathew_loss}
    \end{aligned}
\end{equation}

\begin{figure}[h!]
    \centering
    \begin{subfigure}[t]{0.5\textwidth}
        \centering
        \includegraphics[width=\textwidth]{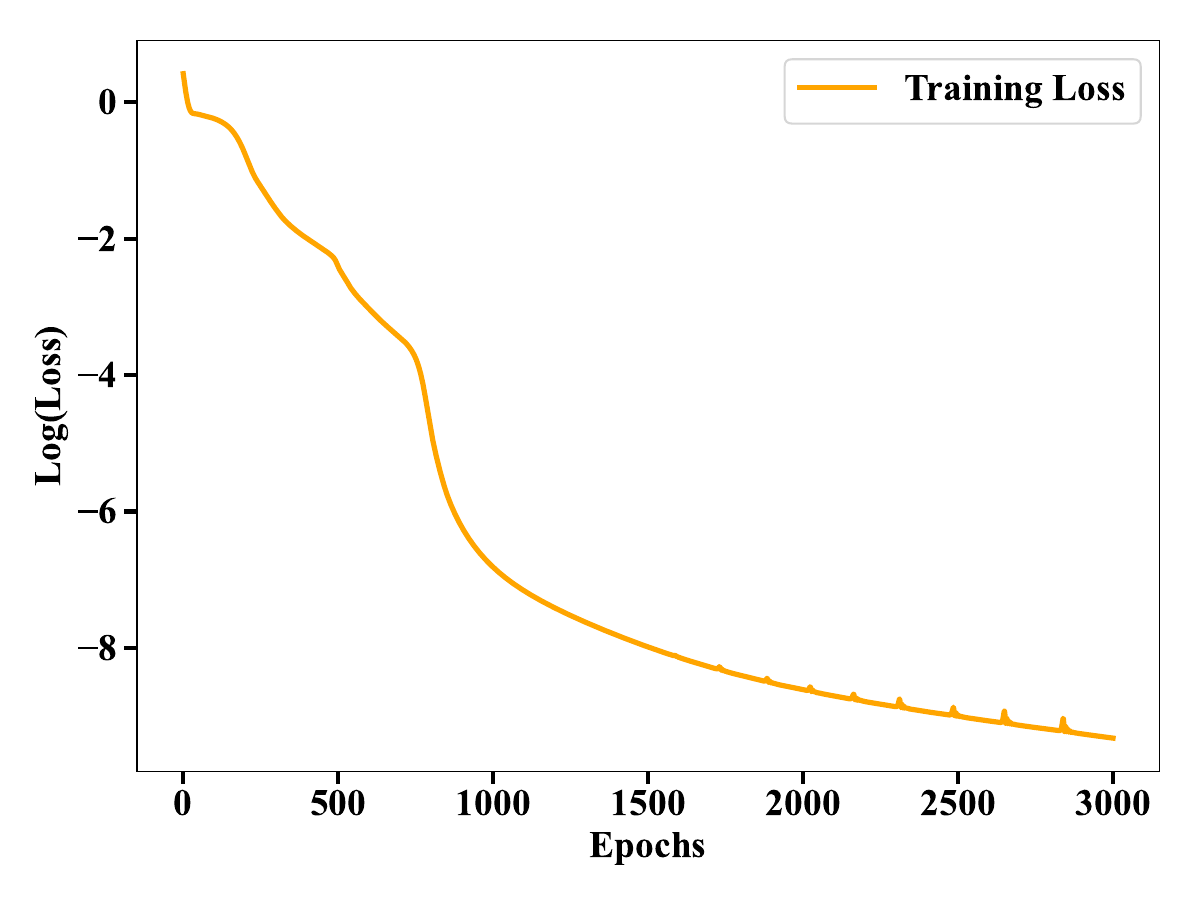}
        \caption{}
        \label{fig-Mathew_case1a}
    \end{subfigure}%
    \hfill
    \begin{subfigure}[t]{0.5\textwidth}
        \centering
        \includegraphics[width=\textwidth]{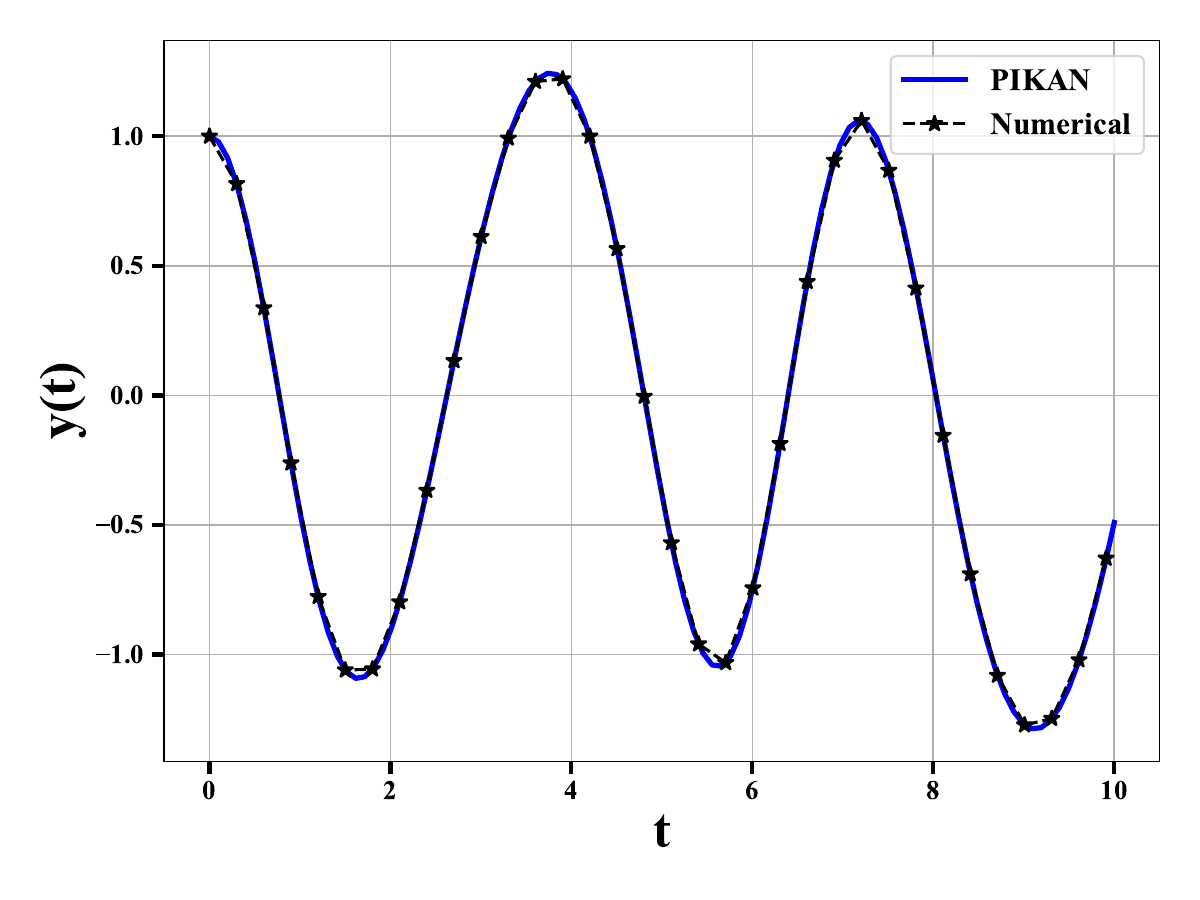}
        \caption{}
        \label{fig-Mathew_case1b}
    \end{subfigure}
    \caption{(a) Illustrates the relationship between loss~\eqref{Mathew_loss} and epoch,(b) Comparison between the PIKAN predicted solution and numerically exact solution of  Eq.~\eqref{Mathew_deqn} for $a =3, \beta = 1.2$.}
\end{figure}
\begin{figure}[h!]
    \centering
    \begin{subfigure}[t]{0.5\textwidth}
        \centering
        \includegraphics[width=\textwidth]{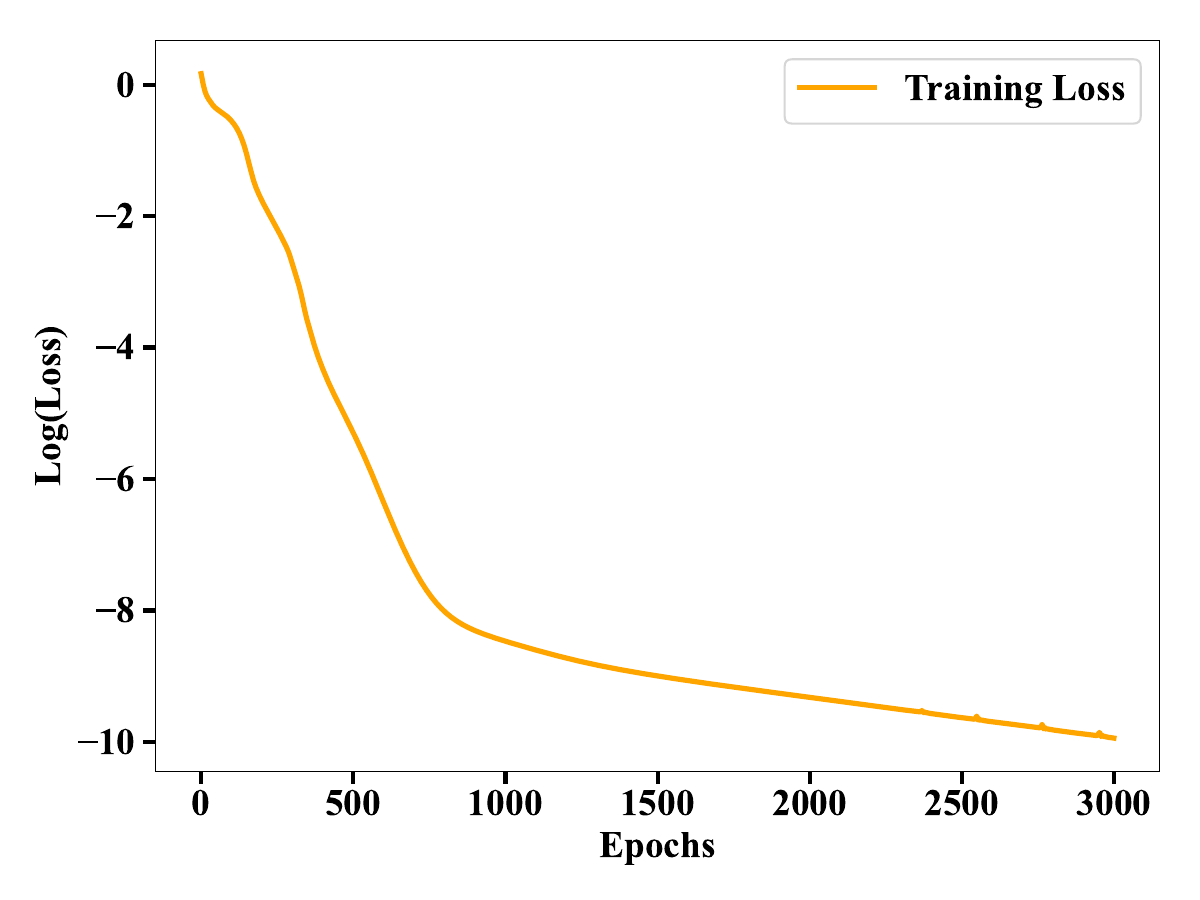}
        \caption{}
        \label{fig-Mathew_case2a}
    \end{subfigure}%
    \hfill
    \begin{subfigure}[t]{0.5\textwidth}
        \centering
        \includegraphics[width=\textwidth]{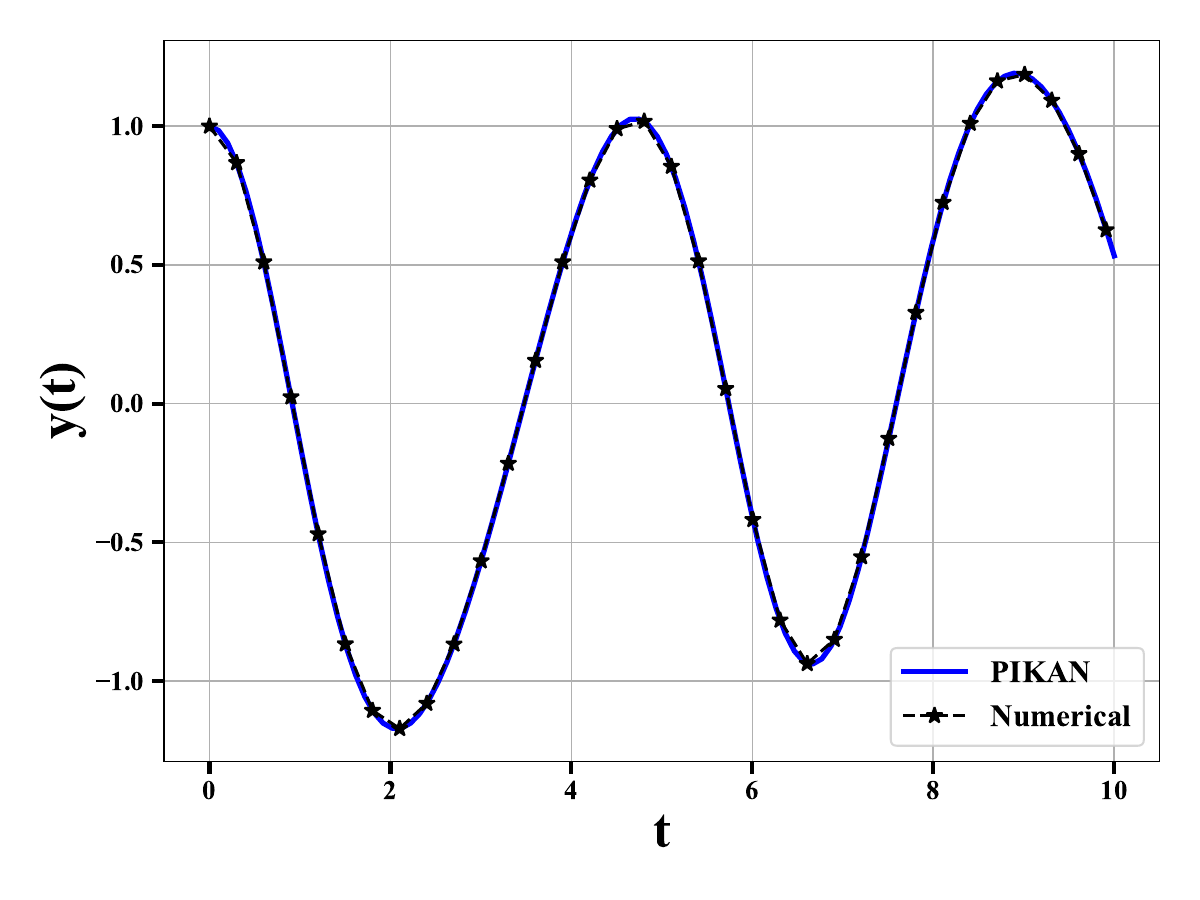}
        \caption{}
        \label{fig-Mathew_case2b}
    \end{subfigure}
    \caption{(a) Illustrates the relationship between loss~\eqref{Mathew_loss} and epoch,(b) Comparison between the PIKAN predicted solution and numerically exact solution of  Eq.~\eqref{Mathew_deqn} for $a =2, \beta = 1 .$}
\end{figure}
\begin{figure}[h!]
    \centering
    \begin{subfigure}[t]{0.5\textwidth}
        \centering
        \includegraphics[width=\textwidth]{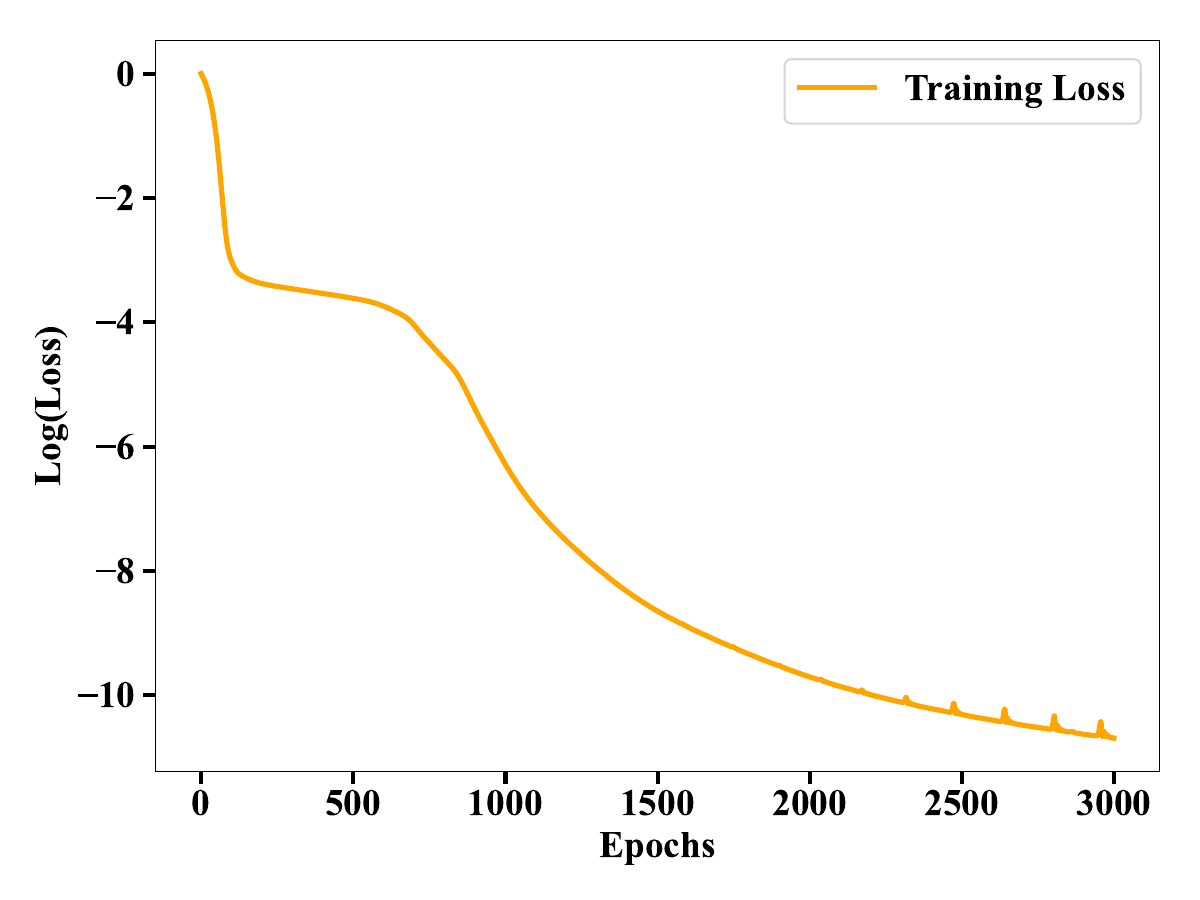}
        \caption{}
        \label{fig-Mathew_case3a}
    \end{subfigure}%
    \hfill
    \begin{subfigure}[t]{0.5\textwidth}
        \centering
        \includegraphics[width=\textwidth]{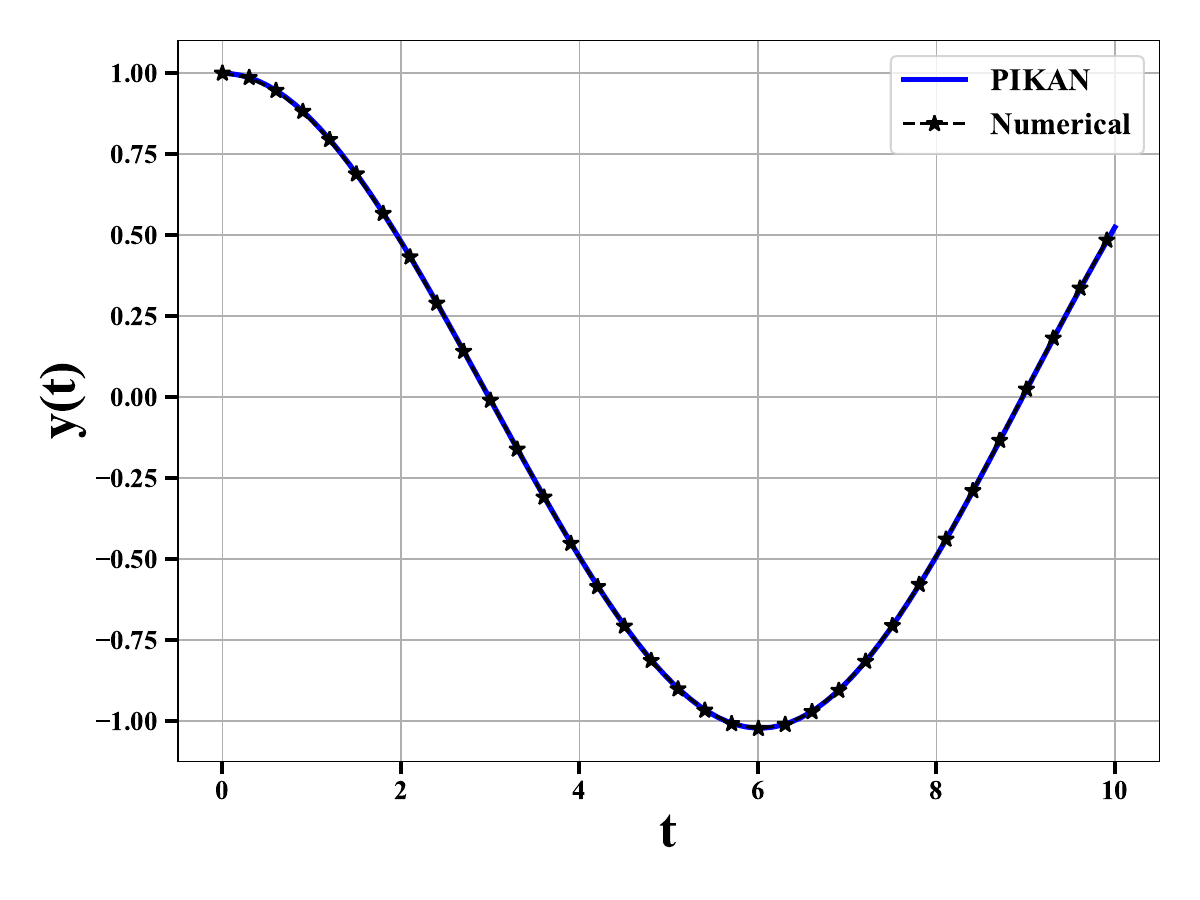}
        \caption{}
        \label{fig-Mathew_case3b}
    \end{subfigure}
    \caption{(a) Illustrates the relationship between loss~\eqref{Mathew_loss} and epoch, (b) Comparison between the PIKAN predicted solution and numerically exact solution of  Eq.~\eqref{Mathew_deqn} for $a =0.25, \beta = 0.05 .$}
\end{figure}
We have utilized WAV-KAN to solve the Mathieu equation~\eqref{Mathew_deqn}
employing  ($N_{r} = 100$) collocation points equally spaced on the interval $0\leq t \leq 10$. The model is trained for $3000$ epochs and we have taken Adam optimizer with a learning-rate $\eta= 0.001$. At the end of $3000$ epochs we have recorded mean squared error loss on the order of $10^{-5}$. The further details of the model is given in Table~\ref{mathew_table}.
The comparison of PIKAN predicted solution and exact numerical solution for different values of the parameters $\alpha$ and $\beta$  are shown in Figs.~\ref{fig-Mathew_case1b},~\ref{fig-Mathew_case2b} and~\ref{fig-Mathew_case3b}.

\begin{table}
\centering
\begin{tabular}{c|c|c}
  \hline
  Architecture  & Wavelet type & Loss \\
  \hline
  \hline
  $[1, 12,8,1]$  & $\sin$ & $10^{-5}$ \\
  \hline
\end{tabular}
\caption{Architecture details of the model used for section~\ref{Mathew_eqn}.}
\label{mathew_table}
\end{table}
\subsection{Van der Pol Equation}
\label{vanderple_eqn}
The Van der Pol equation describes the behavior of a nonlinear oscillator with damping that varies with the amplitude of oscillation. Consider the following initial value problem for Van der Pol equation \cite{Rahman2024}
\begin{equation}
 \label{vanderpol_deqn}
    \begin{aligned}
    y''(t) + \epsilon \left( c_0 + c_1 \cos(\omega t) + \alpha y^2 \right) y'(t) & + \omega_n^2 y(t) \\
    & = f_0 + f_1 \sin(\omega t) 
    \\
    \begin{cases}
    y(0) = 0,
    \\
    y'(0) = 2
    \end{cases}
    \end{aligned}
    %\end{split}
\end{equation}
The parameters are\[
\begin{aligned}
c_0 &= -1, & \epsilon &= 0.2, & \omega_n &= 1, \\
c_1 &= 1, & \alpha &= 1, & \omega &= 0.12, & f_0 &= 0.4
\end{aligned}
\]

We will consider two cases depending on the value of the parameter $f_1=1$ and $f_1=1.7$ respectively.

The residual term is given by
\begin{equation*}
\begin{aligned}
         \mathcal{R}_\theta(t)&= \left[ \hat{y}''(t) + 0.2 \left( -1 + \cos(0.12 t) + \hat{y}(t)^2 \right) \hat{y}'(t) \right. \\
    & \quad \left. + \hat{y}(t) - 0.4 - \sin(0.12 t) \right]^2,   
\end{aligned}
\end{equation*}
and the physics loss, the initial condition loss and the total loss takes the form
\begin{equation}
    %\begin{split}
    \begin{aligned}
    \mathcal{L}_{\mathrm{r}} &=  \frac{1}{N_{r}} \sum_{i=1}^{N_{r}} \left|\mathcal{R}_\theta(t_{\mathrm{r}}^{i})\right |^2, \\
\mathcal{L}_{\mathrm{ic}} &= [\left(\hat{y}(0) - 0 \right)^2 + \left( \hat{y}'(0) - 2 \right)^2],
\\ \qquad\mathcal{L}_{\mathrm{final}}^{DF} &= \mathcal{L}_{\mathrm{r}} + \mathcal{L}_{\mathrm{ic}}
    \label{Loss_VP}
    \end{aligned}
    %\end{split}
\end{equation}

\begin{figure}
    \centering
    \begin{subfigure}[t]{0.5\textwidth}
        \centering
        \includegraphics[width=\textwidth]{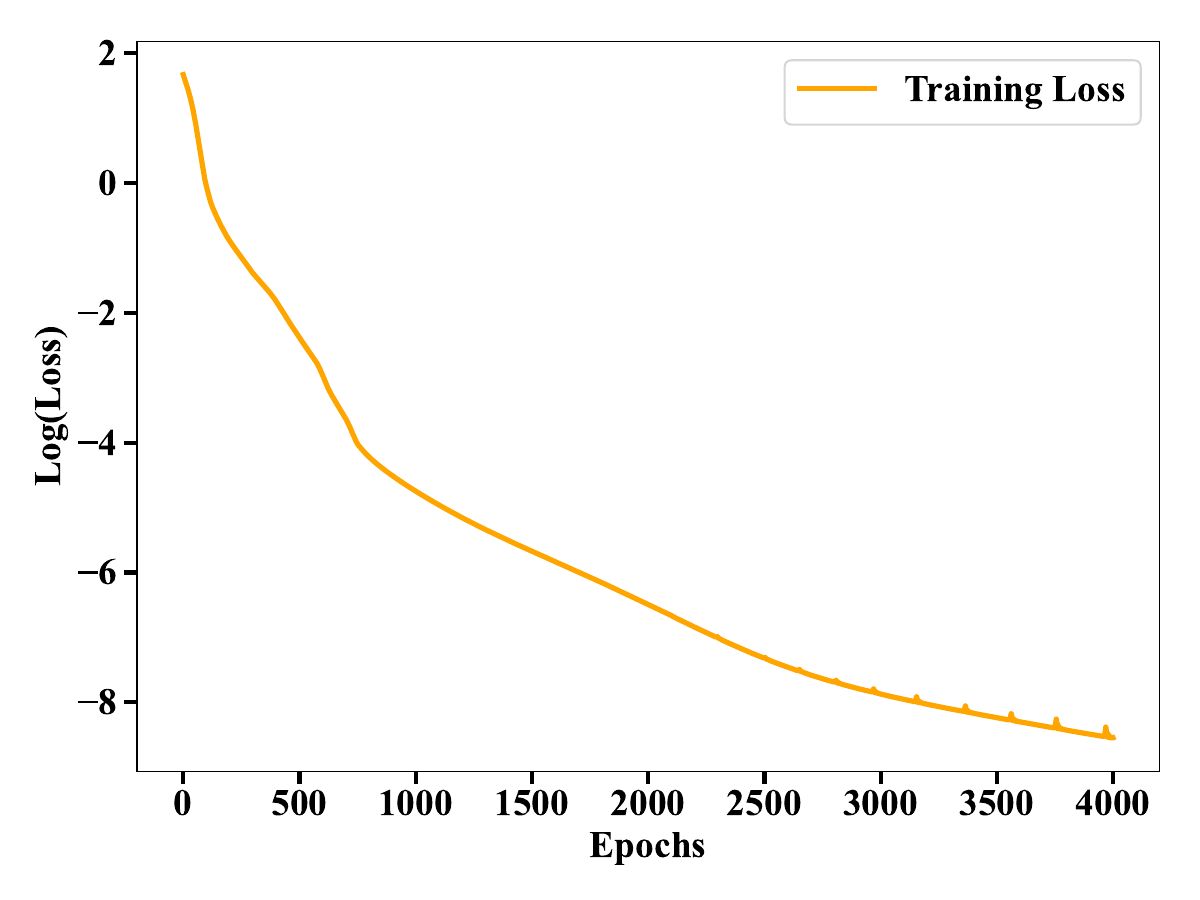}
        \caption{}
        \label{fig-vp_case1a}
    \end{subfigure}%
    \hfill
    \begin{subfigure}[t]{0.5\textwidth}
        \centering
        \includegraphics[width=\textwidth]{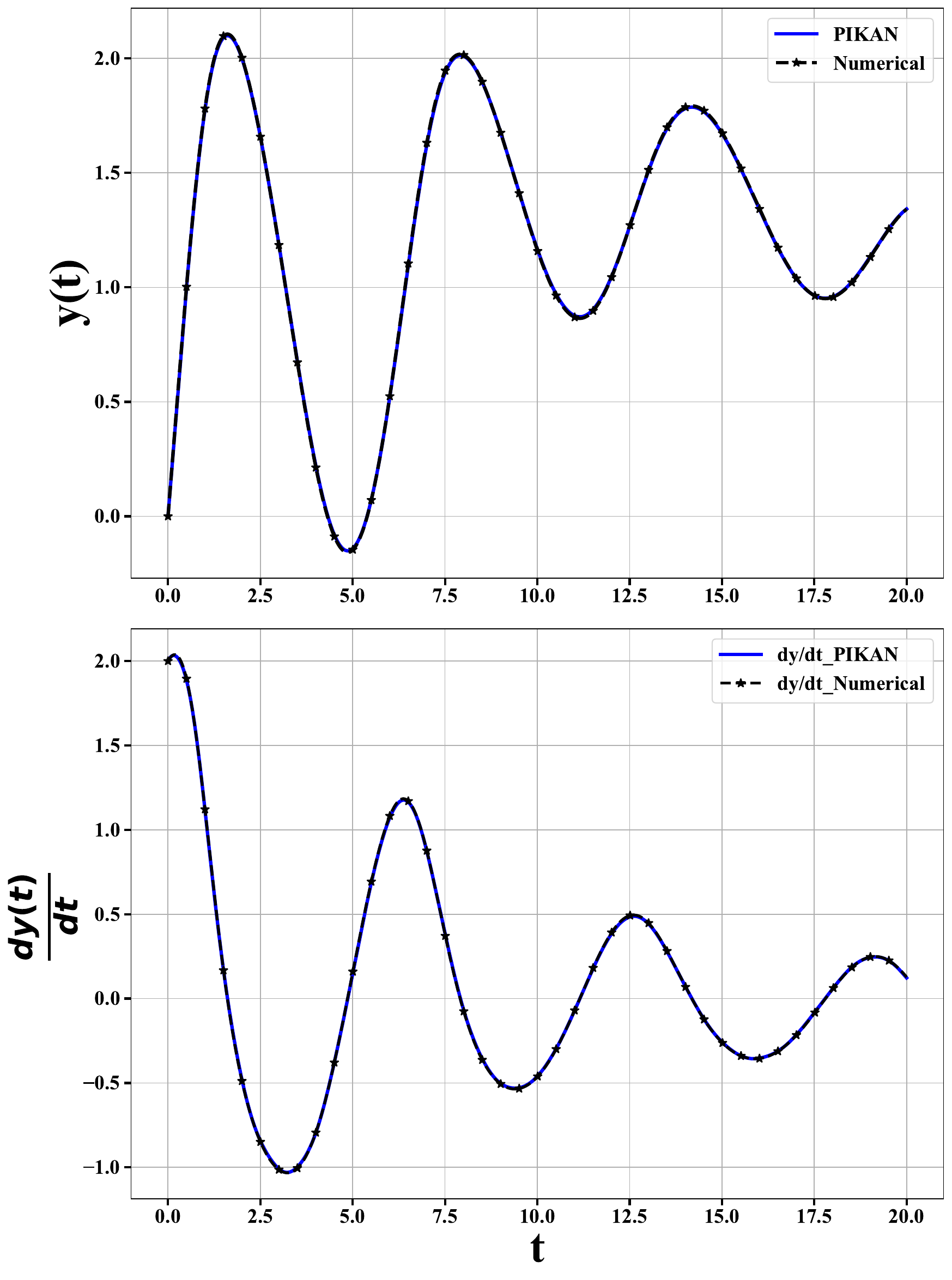}
        \caption{}
        \label{fig-vp_case1b}
    \end{subfigure}
    \caption{(a) Illustrates the relationship between loss~\eqref{Loss_VP} and epoch,(b) Comparison between the PIKAN predicted solution and numerically exact solution of  Eq.~\eqref{vanderpol_deqn} for $f_1 = 1 .$}
\end{figure}

\begin{figure}
    \centering
    \begin{subfigure}[t]{0.45\textwidth}
        \centering
        \includegraphics[width=\textwidth]{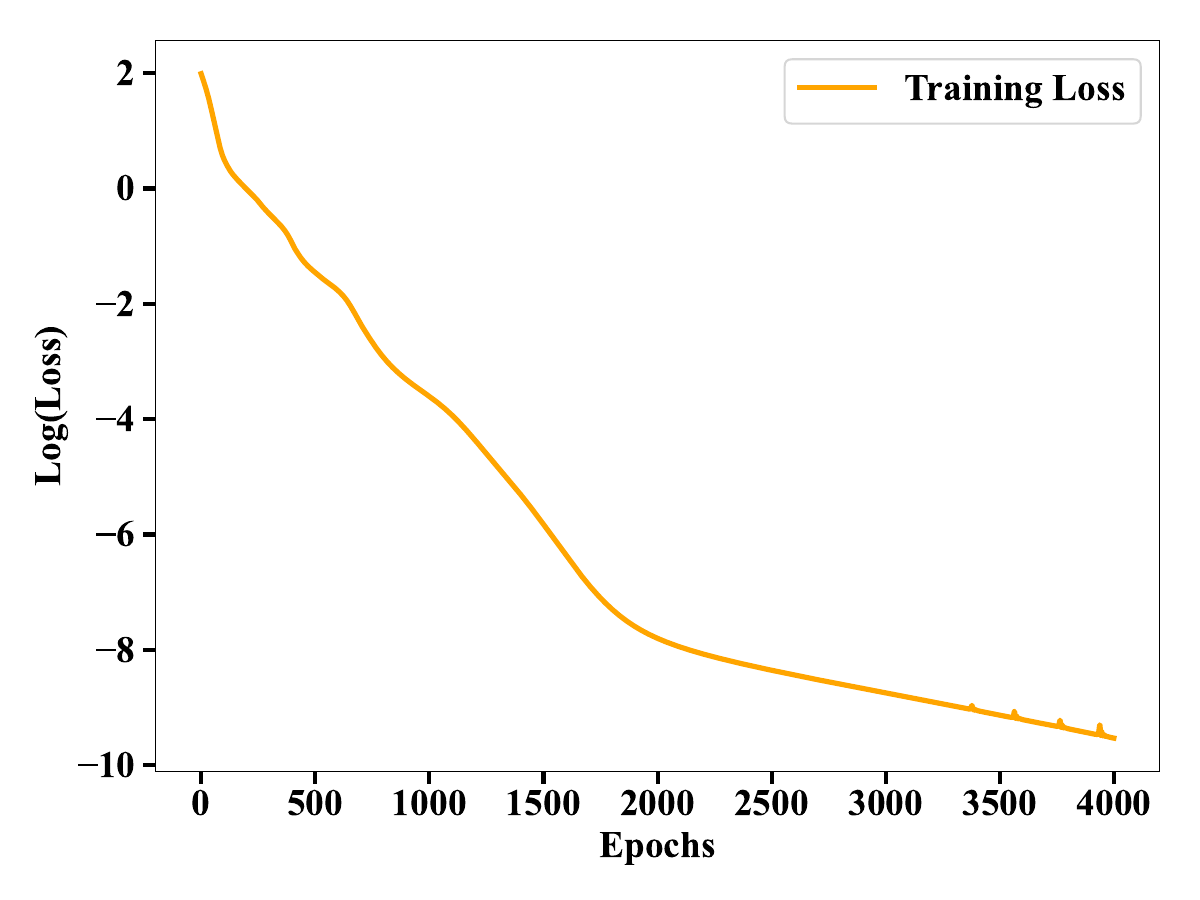}
        \caption{}
        \label{fig-vp_case2a}
    \end{subfigure}%
    \hfill
    \begin{subfigure}[t]{0.45\textwidth}
        \centering
        \includegraphics[width=\textwidth]{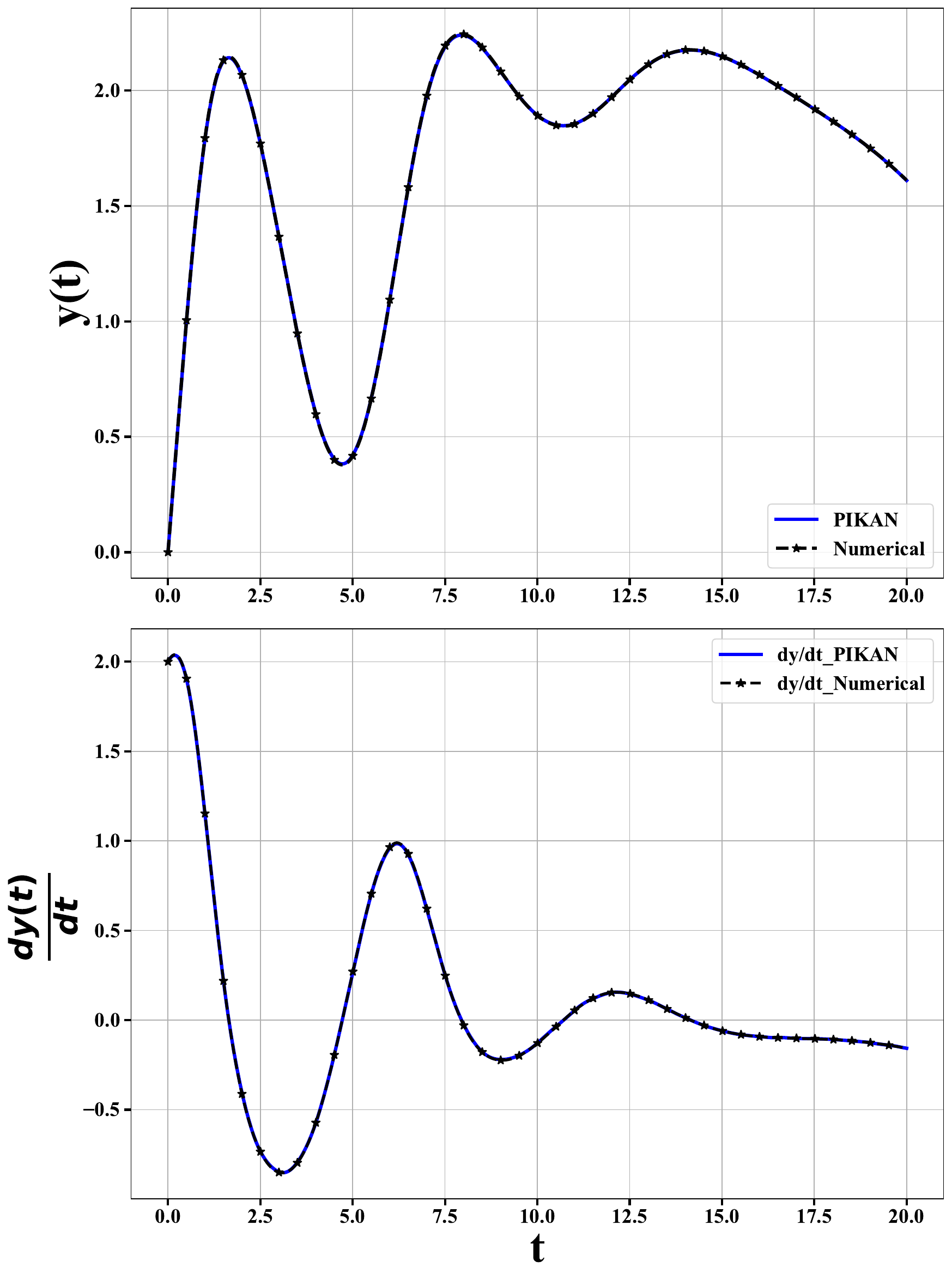}
        \caption{}
        \label{fig-vp_case2b}
    \end{subfigure}
    \caption{(a) Illustrates the relationship between loss~\eqref{Loss_VP} and epoch, (b) Comparison between the PIKAN predicted solution and numerically exact solution of  Eq.~\eqref{vanderpol_deqn} for $f_1 = 1.7$.}
\end{figure}

\begin{table}
\centering
\begin{tabular}{c|c|c}
  \hline
  Architecture  & Wavelet type & Loss \\
  \hline
  \hline
  $[1, 12,8,1]$  & $\sin$ & $10^{-5}$ \\
  \hline
\end{tabular}
\caption{Architecture details of the model used for section~\ref{vanderple_eqn}.}
\label{VP_table}
\end{table}
We have utilized WAV-KAN to solve the Eq~\eqref{vanderpol_deqn} employing $N_t = 100$ collocation points equally spaced on the interval $0 \leq t \leq 20$. The model is trained for 4000 epochs and we have  used Adam optimizer with a variable learning rate $\eta$ starting from $0.001$ and  decreasing $\eta$ by 10 after every 300 epochs. At the end of learning, we have recorded a mean squared loss on the order of $10^{-5}$ (see Fig.~\ref{fig-vp_case1a} and Fig.~\ref{fig-vp_case2a}, respectively). Further details are listed in Table~\ref{VP_table}. The comparison of PIKAN predicted solution and exact numerical solution are given in  Figs.~\ref{fig-vp_case1b} and~\ref{fig-vp_case2b}.

\section{Non-linear partial differential equations} \label{npde}

\subsection{Burgers' Equation}
\label{burger_eqn}

The Burger equation, first introduced by Bateman \cite{Burger0} and later solved by Burger \cite{Burger1}, has a wide range of applications in classical dynamics, nonlinear harmonics, and plasma physics~\cite{Burger2,Burger3,Burger4}. An interesting aspect of this equation is its strong dependence on boundary conditions and the parameter $\nu = 0.1$ [see Eq.~\eqref{burger}]. Depending on the boundary condition, we have divided our investigation of Burgers' equation into two different cases because they require very different approaches: Case 1 can be resolved by minimizing MSE-Loss constructed from training examples, while Case 2 requires more precise training data generated for a specific time scale. This method is known as the data-driven approach~\cite{PINN1}.

Consider Burger's equation with the following boundary and initial condition~\cite{Burgercase12}
\begin{equation}
    \begin{aligned}
    u_t(x,t) + u(x,t) u_x(x,t) = \nu u_{xx}(x,t), \quad
    \begin{cases}
    u(x, 0) = \sin(\pi x), \\
    u(0, t) = 0, \\
    u(1, t) = 0
    \end{cases}
    \end{aligned}
    \label{burger}
\end{equation}
where $x \in [0, 1]$ and $ t \in [0, 1]$.

We can now define the residual term using the PIKAN to-be-learned solution $\hat{u}(x,t)$ as follows
\begin{equation*}
    \mathcal{R}_\theta(x,t) = \hat{u}_t + \hat{u} \hat{u}_x - \nu \hat{u}_{xx}. \nonumber 
\end{equation*}
The physics loss, initial condition loss, boundary loss, data-driven loss and the total loss are the following
%\begin{equation}
\begin{align}
\mathcal{L}_{\mathrm{r}} &=  \frac{1}{N_{r}} \sum_{i=1}^{N_{r}}, |\mathcal{R}_\theta(x_{\mathrm{r}}^{i},t_{\mathrm{r}}^{i})|^2, \nonumber\\ 
\mathcal{L}_{\mathrm{ic}} &= \frac{1}{N_{ic}} \sum_{j=1}^{N_{ic}} \left( \hat{u}(x_{\mathrm{ic}}^{i}, 0) - \sin(\pi x_{\mathrm{ic}}^{j}) \right)^2, \\ \nonumber
\mathcal{L}_{\mathrm{bc}} &= \frac{1}{N_{bc}} \sum_{k=1}^{N_{bc}} \left( \hat{u}(0, t_{\mathrm{bc}}^{k}) \right)^2 + \frac{1}{N_{bc}} \sum_{k=1}^{N_{bc}} \left( \hat{u}(1, t_{\mathrm{bc}}^{k}) \right)^2, \\
\mathcal{L}_{\mathrm{data}}&=\frac{1}{N_{data}} \sum_{p=1}^{N_{data}}\left|\hat{u}\left(x_{\mathrm{data}}^{p}\right)-u\left(x_{\mathrm{data}}^{p}\right)\right|^2,  \nonumber \\ 
\mathcal{L}_{\mathrm{final}}^{DD} &=  \mathcal{L}_{\mathrm{r}} +  \mathcal{L}_{\mathrm{ic}} + \mathcal{L}_{\mathrm{bc}} + \mathcal{L}_{\mathrm{data}}.
\label{burger_loss}
\end{align}
%\end{equation}
where  \(N_{\text{r}}\), \(N_{\text{ic}}\), and \(N_{\text{bc}}\) are the number of points used to calculate the  residual, initial condition loss, and boundary condition loss, respectively.

\begin{figure}
    \centering
    \begin{subfigure}[t]{0.45\textwidth}
        \centering
        \includegraphics[width=\textwidth]{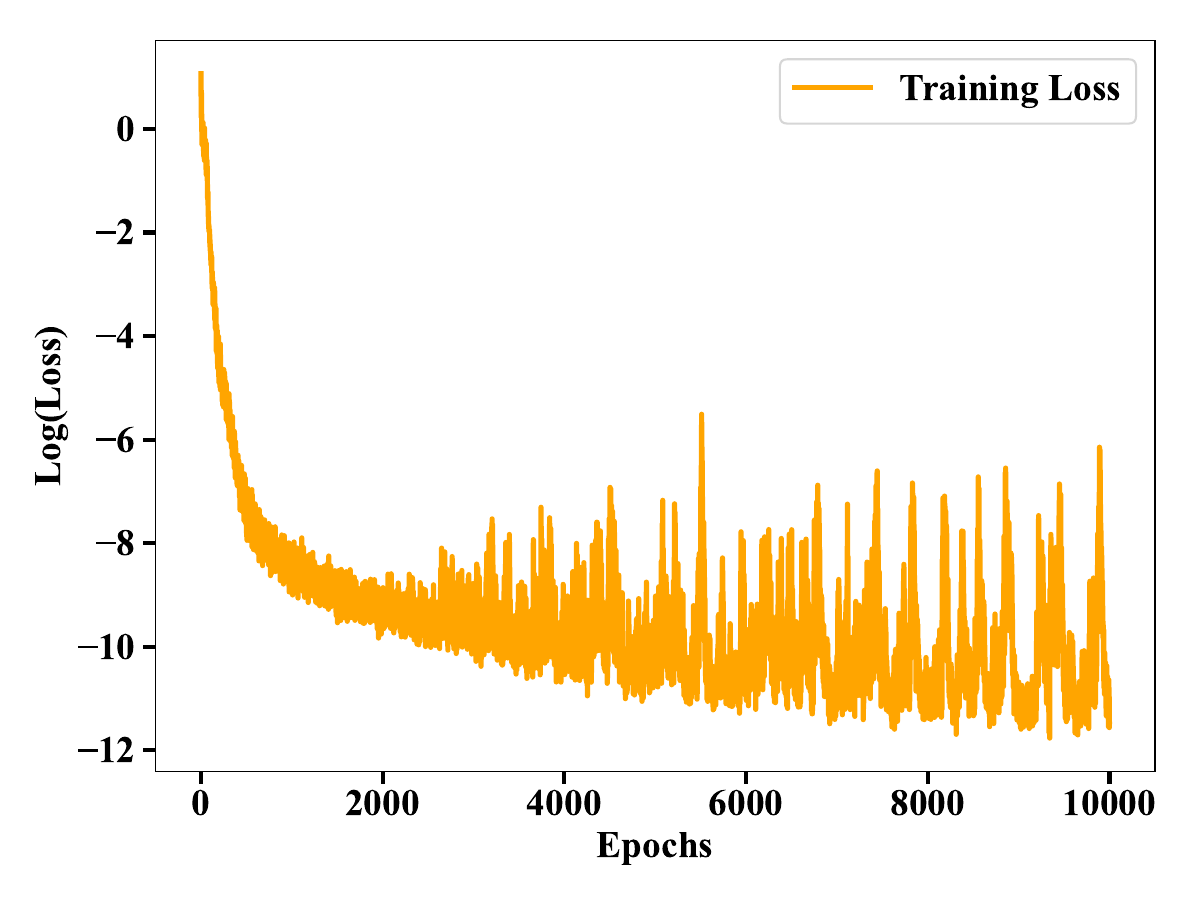}
        \caption{}
        \label{fig-burger_case1a}
    \end{subfigure}%
    \hfill
    \begin{subfigure}[t]{0.45\textwidth}
        \centering
        \includegraphics[width=\textwidth]{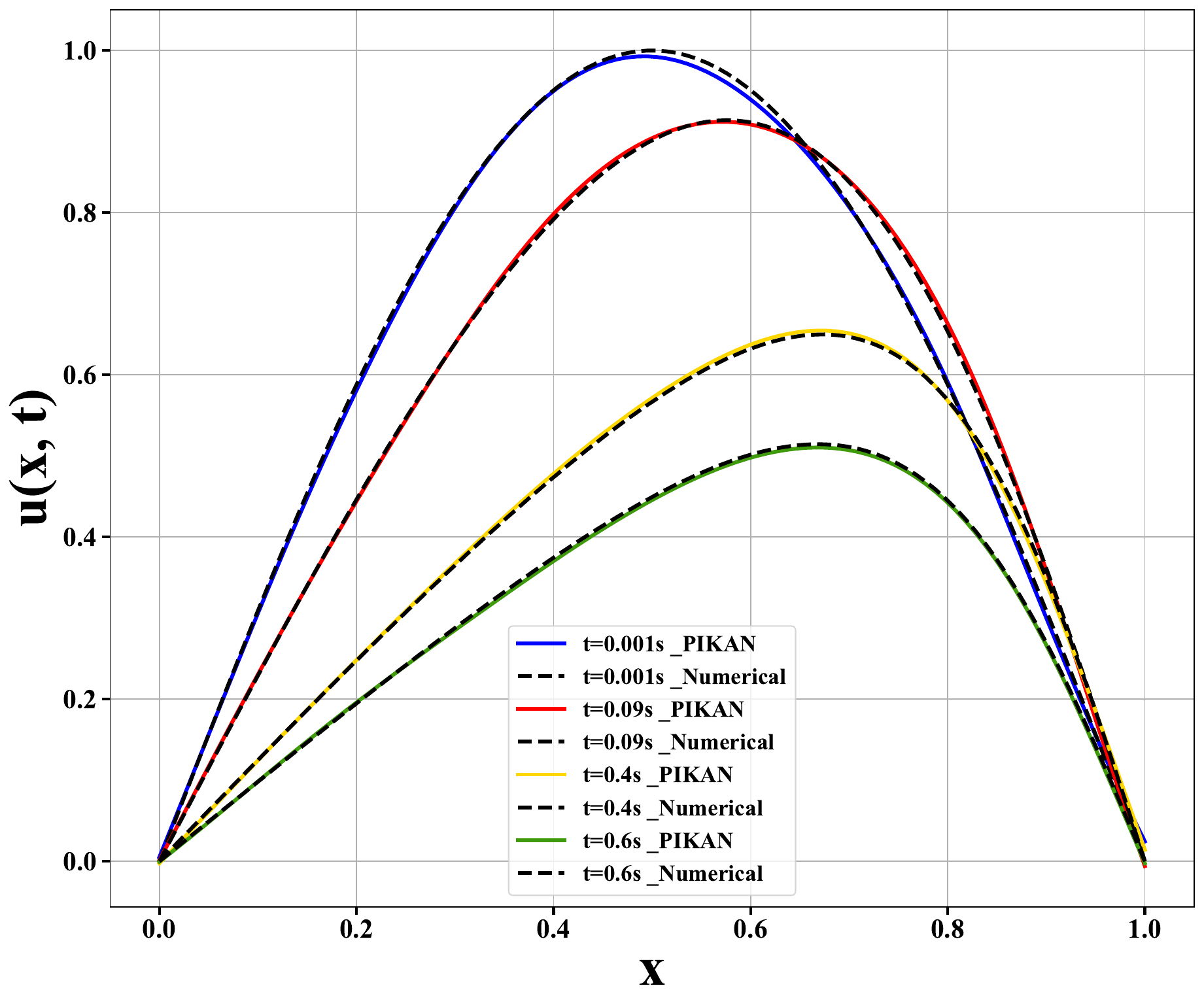}
        \caption{}
        \label{burger_case1b}
    \end{subfigure}
    \caption{(a) Illustrates the relationship between loss~\eqref{burger_loss} and epoch,(b) Comparison between the PIKAN predicted solution and numerically exact solution of  Eq.~\eqref{burger}.}
\end{figure}
The solution of Burgers' Equation~\eqref{burger} is done using the data-driven method~\eqref{weight-loss-DD}.
We have computed the numerical solution first  and then we have taken only 10 \% of the data from the numerical solution and these data points were chosen randomly. Now we have added an extra loss term~\eqref{burger_loss} that corresponds to the data-driven loss [Eqs.~\eqref{burger_loss} and \eqref{weight-loss-DD}]. To train the model we have utilized the PI-efficienKAN by employing $N_{x} = 100$ and $N_{t} = 100$ collocation points distributed uniformly over the interval $0\leq x\leq 1$ and $0\leq t \leq 1$. The model is trained for $10^4$ epochs and to minimize the loss function we have used the AdamW optimizer with a variable learning rate $\eta$ starting from $0.005$ and  decreasing $\eta$ by a factor of 0.1 after every $10^3$ epochs. At the end of training, we have recorded a mean squared error loss on the order of $10^{-5}$ (see Fig.~\ref{fig-burger_case1a}).
Further details of the model is listed in Table~\ref{Burger_table}.
The comparison of PIKAN predicted solution and the numerically exact solution of Eq.~\eqref{burger} computed for different time steps is shown in Fig.~\ref{burger_case1b}.

\begin{table}
\centering
\normalsize
\resizebox{\linewidth}{!}{
\begin{tabular}{c|c|c|c|c}
  \hline
  \textbf{Architecture} & \textbf{Spline-order} & \textbf{Grid size} & \textbf{Basis Activation} & \textbf{Loss} \\
  \hline
  \hline
  $[2,8,4,1]$ & 3 & 5 & $\sin$ & $10^{-5}$ \\
  \hline
\end{tabular}}
\caption{The architecture details for Sec.~\ref{burger_eqn}.}
\label{Burger_table}
\end{table}

\subsection{Allen-Cohen Equation}
\subsubsection{Case - 1}
\label{Allen_case1}

The Allen-Cohen equation serves as a valuable model in materials science for elucidating phase separation and interface dynamics in multi-component alloy systems. This particular reaction-diffusion equation is employed to depict the progression of an order parameter that distinguishes between different phases of a material. In this section, we will delve into solutions for this highly nonlinear system using PIKAN \cite{allen-cohen_2}.

The Allen-Cohen equation is provided below, and for this particular equation, we have selected a value of $\nu = 0.001$,

\begin{align}
    \label{allen_1}
    u_t &= \nu \, u_{xx} - u^3 + u , \notag\\
    &\begin{cases}
    u(x, 0) &= 0.53x + 0.47\sin(-1.5\pi x) \\
    u(1, t) &= 1 \\
    u(-1, t) &= -1,
    \end{cases} 
\end{align}
where $x \in [-1, 1]$ and $ t \in [0, 1]$.

The residual, physics loss, initial condition loss, boundary condition loss, and total loss are the following:
\begin{align}
\mathcal{R}_\theta(x,t) &= \hat{u_t} + \hat{u}^{3} - \hat{u} - \nu \hat{u}_{xx},  \nonumber \\
\mathcal{L}_{\mathrm{r}} &=  \frac{1}{N_{{r}}} \sum_{i=1}^{N_{{r}}} |\mathcal{R}_\theta(x_{\mathrm{r}}^{i},t_{\mathrm{r}}^{i})|^2, \nonumber \\
\mathcal{L}_{\mathrm{ic}} &= \frac{1}{N_{{ic}}} \sum_{j=1}^{N_{{ic}}} (\hat{u}(x_{\mathrm{ic}}^{j}, 0) - u(x_{\mathrm{ic}}^{j},0))^2, \nonumber \\
\mathcal{L}_{{bc}} &= \frac{1}{N_{{bc}}} \sum_{k=1}^{N_{{bc}}} \left[ (\hat{u}(1, t_{\mathrm{bc}}^{k}) - 1)^2 + (\hat{u}(-1, t_{\mathrm{bc}}^{k}) + 1)^2 \right], \nonumber \\
\mathcal{L}_{\mathrm{final}}^{DF} &= \mathcal{L}_{\mathrm{r}} + \mathcal{L}_{\mathrm{ic}} + \mathcal{L}_{\mathrm{bc}}. \label{eq:final_loss_allen_case1}
\end{align}

\begin{figure}
    \centering
    \begin{subfigure}[t]{0.48\textwidth}
        \centering
        \includegraphics[width=\textwidth]{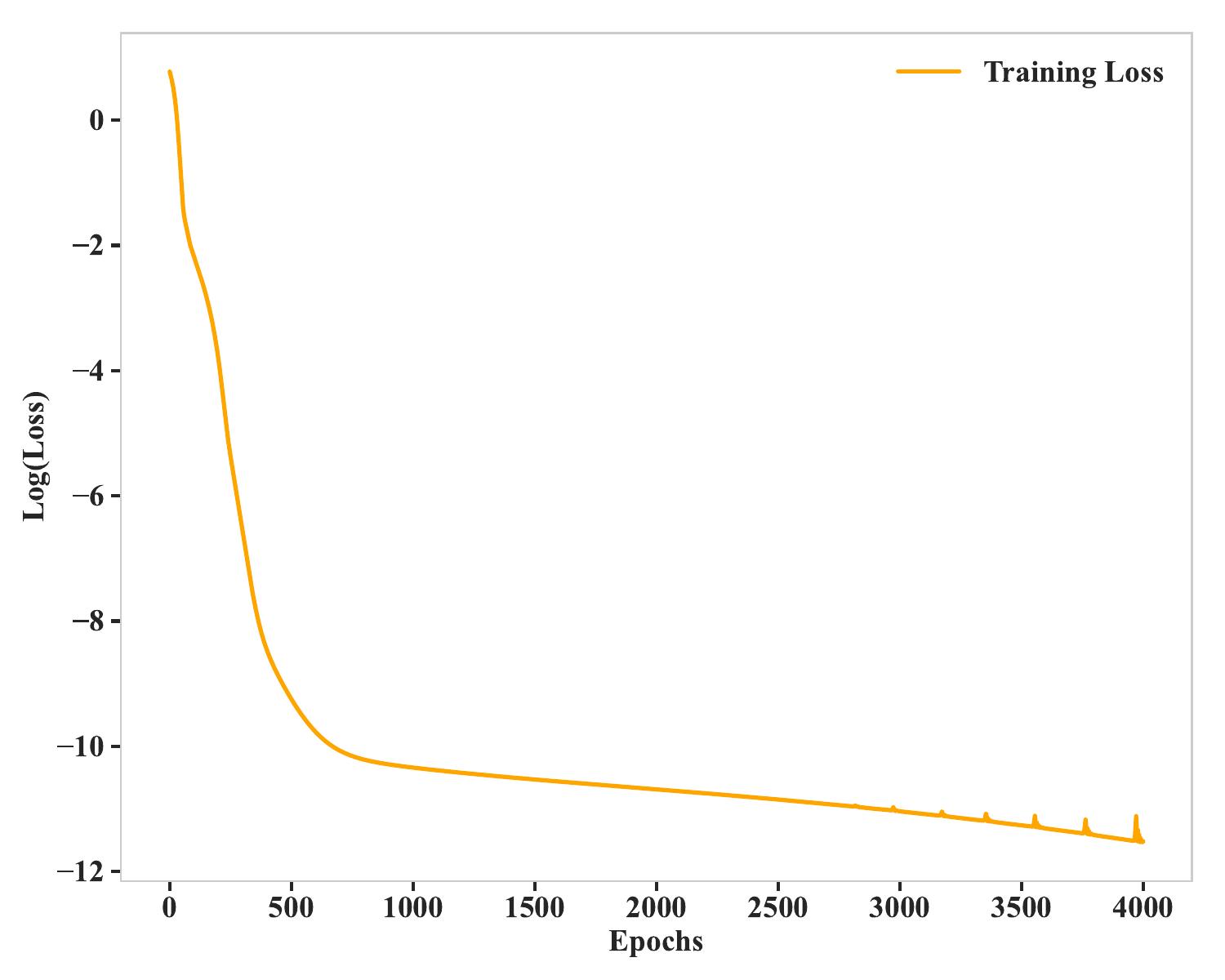}
        \caption{}
        \label{fig-allen_case1a}
    \end{subfigure}%
    \hfill
    \begin{subfigure}[t]{0.48\textwidth}
        \centering
        \includegraphics[width=\textwidth]{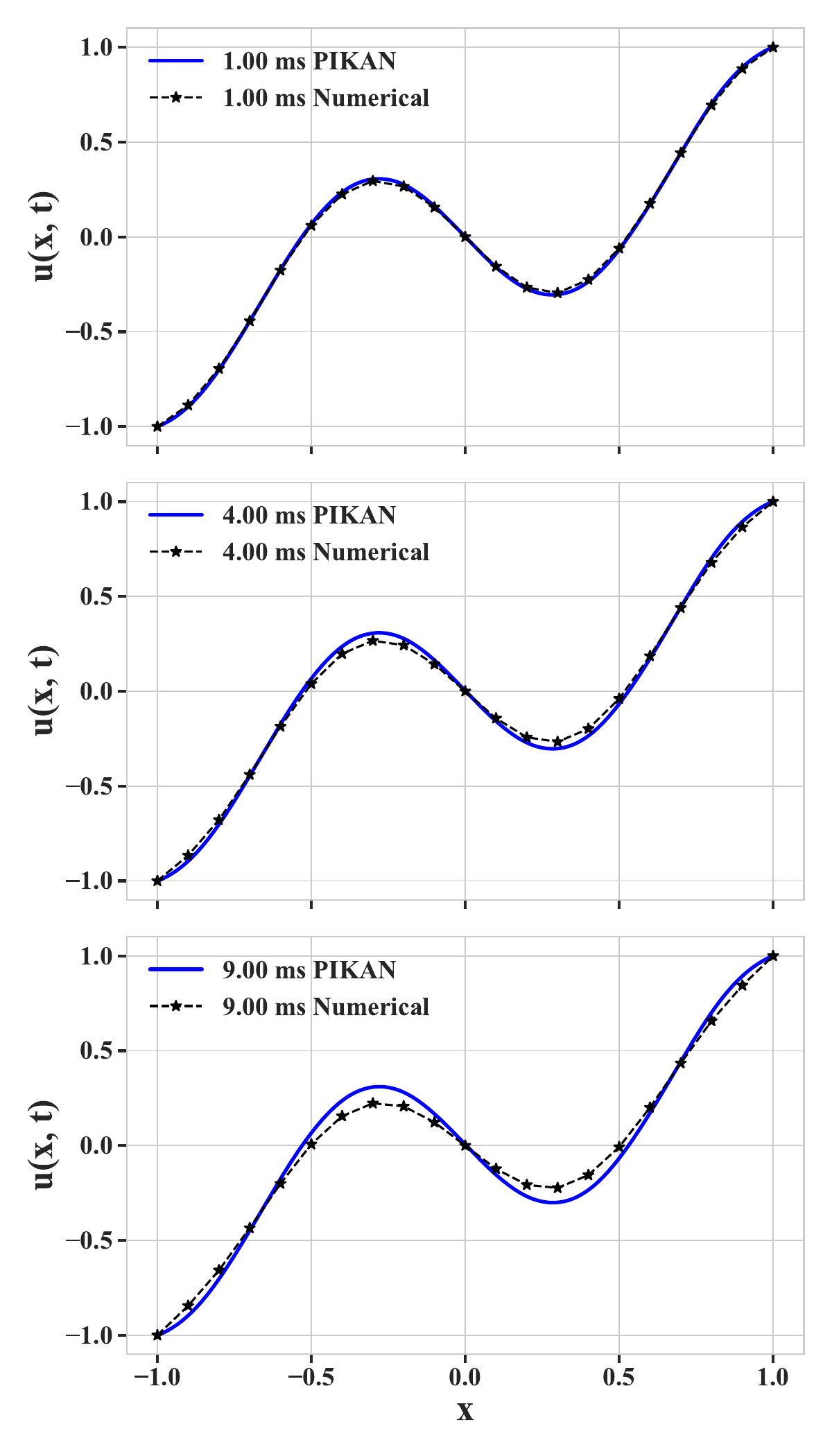}
        \caption{}
        \label{allen_case1b}
    \end{subfigure}
    \caption{(a) Illustrates the relationship between loss~\eqref{eq:final_loss_allen_case1} and epoch,(b) Comparison between the PIKAN predicted solution and numerically exact solution of  Eq.~\eqref{allen_1}.}
\end{figure}

We have utilized efficient-KAN to solve Eq.~\eqref{allen_1}, employing  $N_{x} = 100$ collocation points for space and $N_{t} = 100$ for time, uniformly spread over the interval $-1\leq x \leq 1$ and  $0\leq t \leq 1 $. The model is trained for 4000 epochs and to  minimize the loss function we have used the AdamW optimizer with a learning-rate $\eta = 0.001$. At the end of learning, we have recorded the MSE loss on the order of $10^{-6}$. Further details of the model is listed in Table~\ref{allen_table}.
The comparison between the PI-KAN predicted solution and the numerically exact solution of Eq.~\eqref{allen_1} for different time is shown in Fig.~\ref{allen_case1b}
\cite{allen-cohen_2}.

\begin{table}
\centering
\begin{tabular}{c|c|c|c|c}
  \hline
  Architecture & Spline-order & Grid size & Basis Activation & Loss \\
  \hline
  \hline
  $[2,12,8,12,1]$ & 3 & 5 & $\sin$ & $10^{-6}$ \\
  \hline
\end{tabular}
\caption{Architecture details of the model used in section~\ref{Allen_case1}}
\label{allen_table}
\end{table}

\subsubsection{Case - 2}
\label{allen_case2}

Now, we will solve a slightly more challenging version of the Allen-Cohen equation  \cite{allen-cohen}
\begin{align}
    \label{allen_2}
    u_t = \nu u_{xx} - 5 u^3 + 5 u,
    \begin{cases}
    u(x, 0)&= x^{2}\cos(\pi x)\\
    u(1, t) &=1 \\
    u(-1, t) &=-1,
    \end{cases} 
\end{align}
where $\nu = 0.0001$, $x \in [-1, 1]$, and $t \in [0, 1]$.

The losses are
\begin{align}
\begin{split}
\mathcal{R}_\theta(x,t) &= \hat{u}_t + 5 \hat{u}^3 - 5 \hat{u} - \nu \hat{u}_{xx}, \\
\mathcal{L}_{\mathrm{r}} &=  \frac{1}{N_{r}} \sum_{i=1}^{N_{r}} |\mathcal{R}_\theta(x_{\mathrm{r}}^{i},t_{\mathrm{r}}^{i})|^2, \\
\mathcal{L}_{\mathrm{ic}} &= \frac{1}{N_{ic}} \sum_{j=1}^{N_{ic}} (\hat{u}(x_{\mathrm{ic}}^{j}, 0) -(x_{\mathrm{ic}}^{j})^{2}\cos(\pi x_{\mathrm{ic}}^{j}))^2, \\
\mathcal{L}_{\mathrm{bc}} &= \frac{1}{N_{bc}} \sum_{k=1}^{N_{bc}} \left[ (\hat{u}(1, t_{\mathrm{bc}}^{k}) - 1)^2 + (\hat{u}(-1, t_{\mathrm{bc}}^{k}) + 1)^2 \right], \\
\mathcal{L}_{\mathrm{data}} &= \frac{1}{N_{data}} \sum_{p=1}^{N_{data}} \left|\hat{u}(x_{\mathrm{data}}^{p})-u(x_{\mathrm{data}}^{p})\right|^2, \\
\end{split} \nonumber \\
\mathcal{L}_{\mathrm{final}}^{DD} &= \mathcal{L}_{\mathrm{r}} + \mathcal{L}_{\mathrm{ic}} + \mathcal{L}_{\mathrm{bc}} + \mathcal{L}_{\mathrm{data}}. \label{allen_loss2}
\end{align}

\begin{figure}[h!]
    \centering
    \begin{subfigure}[t]{0.5\textwidth}
        \centering
        \includegraphics[width=\textwidth]{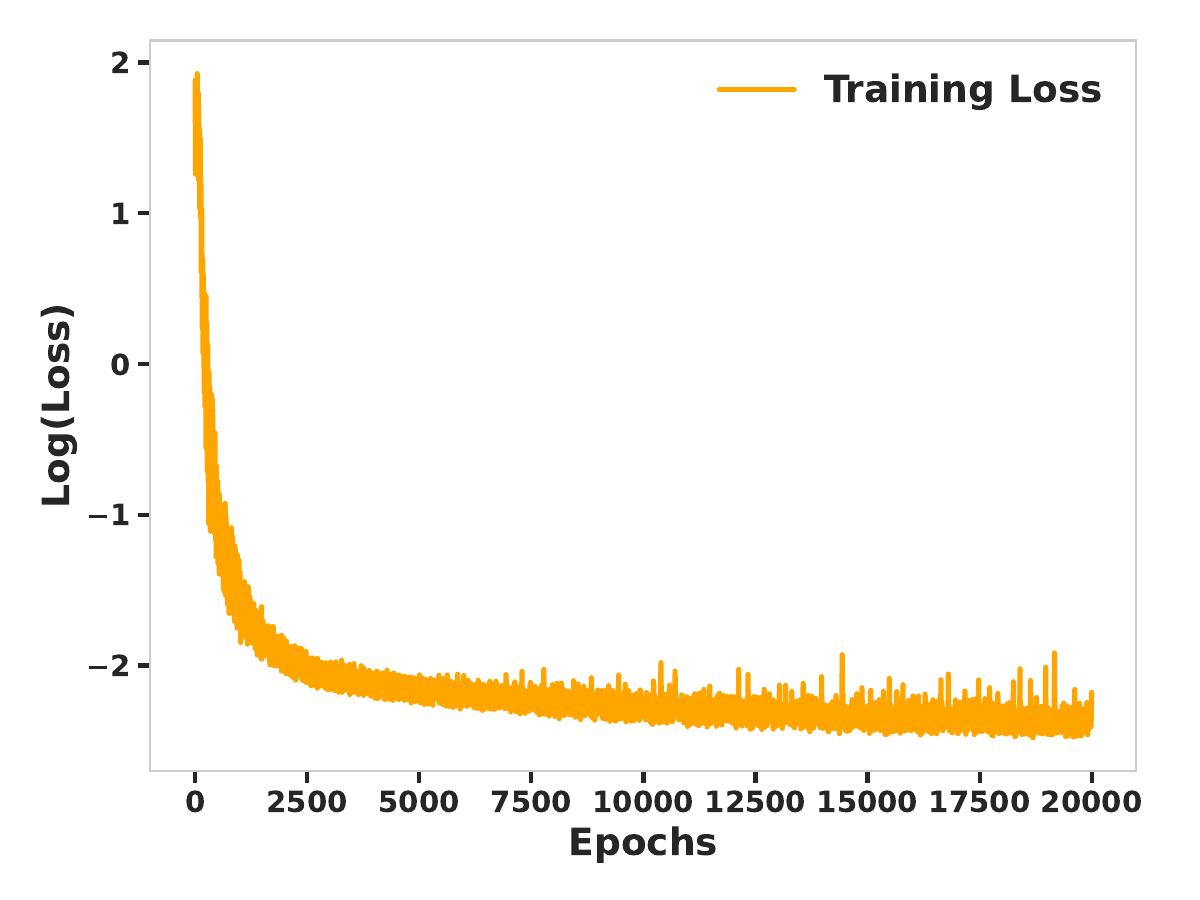}
        \caption{}
        \label{fig-allen_case2a}
    \end{subfigure}%
    \hfill
    \begin{subfigure}[t]{0.5\textwidth}
        \centering
        \includegraphics[width=\textwidth]{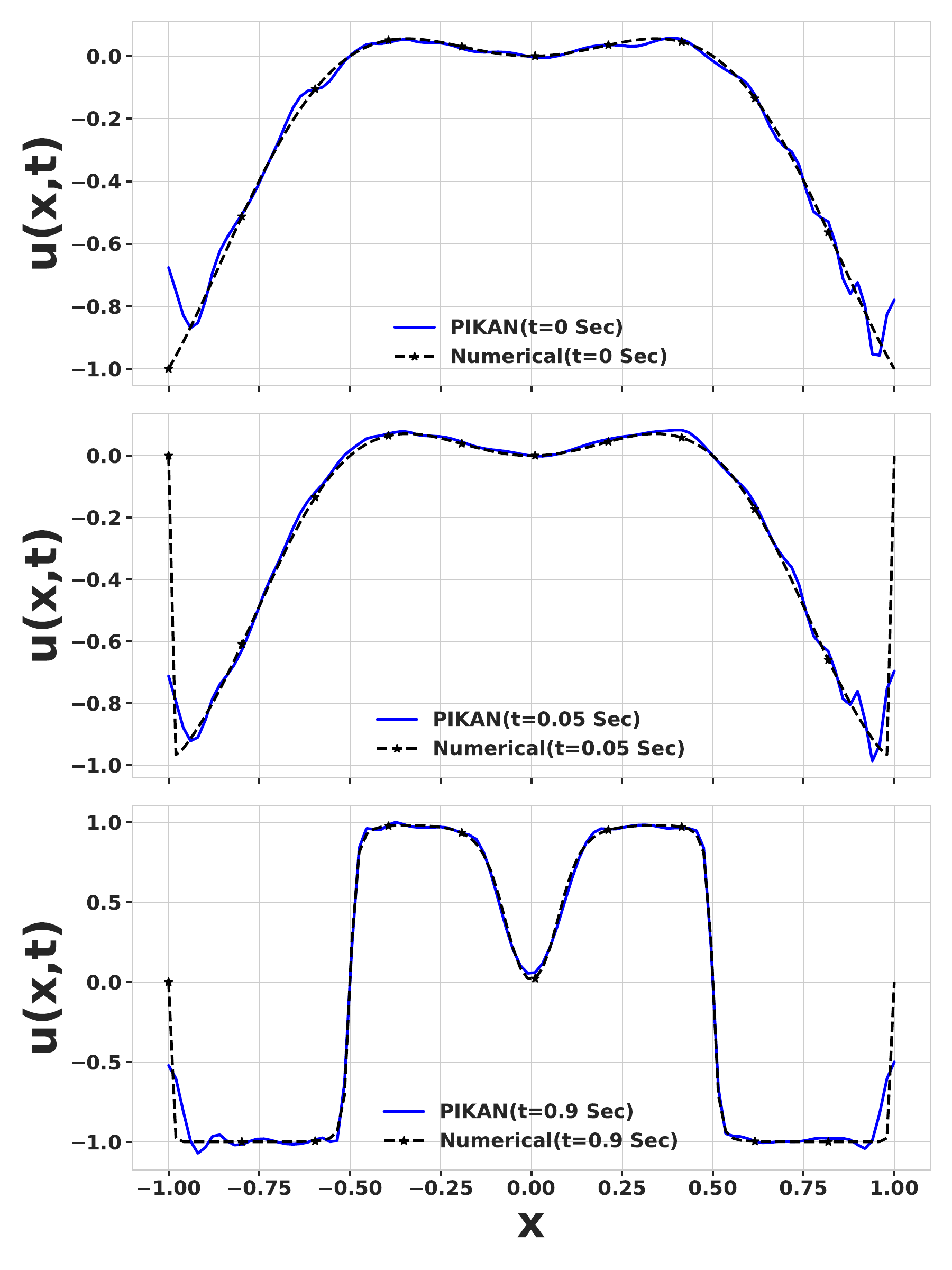}
        \caption{}
        \label{allen_case2b}
    \end{subfigure}
    \caption{(a) Illustrates the relationship between loss~\eqref{allen_loss2} and epoch,(b) Comparison between the PIKAN predicted solution and numerically exact solution of  Eq.~\eqref{allen_2}.}
\end{figure}
\begin{table}
\centering
\begin{tabular}{|c|c|c|c|c|}
  \hline
  Architecture & Spline-order & Grid size & Basis Activation & Loss \\
  \hline
  \hline
  $[2,8,6,1]$ & 3 & 5 & $\sin$ & $0.08$ \\
  \hline
\end{tabular}
\caption{Architecture details of the model used in section~\ref{allen_case2}.}
\label{Allen_case2_table}
\end{table}

The solution of the Allen-Cohen equation~\eqref{allen_2} is obtained using the data-driven method~\eqref{weight-loss-DD}. Initially, the numerical solution is computed, and then only 10\% of the data from the numerical solution is randomly selected. An additional loss term~\eqref{allen_loss2} has been included in the loss function, corresponding to the Data-driven loss [Eqs.~\eqref{allen_loss2} and \eqref{weight-loss-DD}]. To train the model, we have utilized the efficient-KAN with $N_{x} = 100$ and $N_{t} = 100$ collocation points uniformly distributed over the interval $-1\leq x\leq 1$ and $0\leq t \leq 1$. The model is trained for 20000 epochs, using the AdamW optimizer with a learning rate of $\eta = 0.001$ to minimize the loss function.  At the end of traning, a mean squared error loss on the order of $10^{-2}$ is recorded (see Fig.~\ref{fig-allen_case2a}). Additional details of the model are provided in Table~\ref{Allen_case2_table}. A comparison of the PIKAN predicted solution and the numerically exact solution of Eq.~\eqref{allen_2} for different time steps is shown in Fig.~\ref{allen_case2b}.

\section{Comparing Physics-Informed Models: PINN and PIKAN} \label{diss}
The PIKAN method marks a step forward in solving differential equations, offering improvements over the traditional PINN approach. A major advantage of PIKAN is its simplified hyperparameter tuning compared to PINNs. Further, PIKAN stands out for its faster convergence rates, enabling quicker and more reliable solutions. This is a crucial benefit, as it means complex differential equations can be tackled more efficiently and with higher accuracy.

Before diving into the comparative analysis, it is crucial to understand the PIKAN architecture, which can be represented as a graph, where nodes denote summation operation and edges -- learnable activation functions.  Consider, as an example, the KAN architecture of form $[\textbf{I},\textbf{a},\textbf{b},\textbf{O}]$, where \textbf{I} denotes the number of input variables, \textbf{a} represents the number of nodes in the first hidden layer, \textbf{b} -- the number of nodes in the second hidden layer, and \textbf{O} -- the number of output nodes. 

Let us now explore how PIKAN simplifies hyperparameter tuning compared to PINNs. First, for the case of the system of two linear equations~\eqref{LCDE_17}, two separate neural networks have been trained in~\cite{cde1}: The PINN model approximating $u(x)$ [Eq.~\eqref{LCDE_17}] has 20 neurons and 4 hidden layers, and the PINN model for $v(x)$ has 10 neurons and 5 hidden layers. A single PIKAN model presented in Section~\ref{Linear_Coupled_single_Variable} with architecture $[1,2,3,2]$ (recall that there is only two intermediate layers) already performed well. For the system of equations represented by \eqref{Nonlinear_Deqn}, the PINN model required 8 hidden layers with 8 neurons each, as detailed in \cite{cde1}. In contrast, the corresponding PIKAN model (in Section~\ref{Nonlinear_Coupled_diffferentialequation_sollution_technique}) achieved similar accuracy with just a single layer having an architecture of $[1, 7, 2]$.

For the Van Der Pol equation \eqref{vanderpol_deqn}, in \cite{Rahman2024}, a PINN network  has been constructed with 3 hidden layers, each with 64 neurons; whereas, in the PIKAN case (in Section \ref{vanderple_eqn}), a simple architecture $[1,12,8,1]$ works. Further, a neural network with 4 hidden layers and 200 neurons each was used in \cite{PINN1} to solve Allen Cohen’s system~\eqref{allen_2}. In comparison, our PIKAN implementation (Section~\ref{allen_case2}) requires a simple architecture $[2,8,6,1]$
containing only 2 hidden layers.

To illustrate faster convergence in PIKAN, note that when solving the Burgers' equation (Section~\ref{burger_eqn}), we observed that PINN method resulted in a mean squared error loss of $10^{-3}$~\cite{cde1}, whereas PIKAN achieved the loss of $10^{-5}$, after training both methods for 5000 epochs.

\section{Conclusion} \label{concl}

We have demonstrated that physics-informed modeling using KAN via efficient-KAN and wave-KAN are versatile and efficient tools. We showcased their performance on a series of differential equations using data-free and data-driven approaches and conducted a comparative analysis in terms of architectural complexity and performance with PINN.

\begin{acknowledgments}
D.I.B and A.S. are supported by Army Research Office (ARO) (grant W911NF-23-1-0288; program manager Dr.~James Joseph). The views and conclusions contained in this document are those of the authors and should not be interpreted as representing the official policies, either expressed or implied, of ARO or the U.S. Government. The U.S. Government is authorized to reproduce and distribute reprints for Government purposes notwithstanding any copyright notation herein.
\end{acknowledgments}

\bibliographystyle{apsrev4-1} % Tell bibtex which bibliography style to use
\bibliography{References.bib} % Tell bibtex which .bib file to use (this one is some example file in TexLive's file tree)

\end{document}